\documentclass[sigconf,authorversion,nonacm]{acmart}

\usepackage{booktabs} 
\usepackage{graphicx}
\usepackage{subcaption}
\usepackage{multirow}
\usepackage{orcidlink}

\begin{document}
\title{Enhancing Traffic Flow Prediction using Outlier-Weighted AutoEncoders: Handling Real-Time Changes
}

\author{Himanshu Choudhary~\orcidlink{0009-0006-2779-3049}}
\affiliation{%
  \institution{Department of Mathematics and Computer Science Eindhoven University of Technology}
  \city{Eindhoven} 
\country{Netherlands}
}
\email{himanshu.dce12@gmail.com}

\author{Marwan Hassani~\orcidlink{0000-0002-4027-4351}}

\affiliation{%
  \institution{Department of Mathematics and Computer Science Eindhoven University of Technology}
    \city{Eindhoven} 
\country{Netherlands}
 }
\email{m.hassani@tue.nl}

\begin{abstract}
In today's urban landscape, traffic congestion poses a critical challenge, especially during outlier scenarios. These outliers can indicate abrupt traffic peaks, drops, or irregular trends, often arising from factors such as accidents, events, or roadwork. Moreover, Given the dynamic nature of traffic, the need for real-time traffic modeling also becomes crucial to ensure accurate and up-to-date traffic predictions. To address these challenges, we introduce the Outlier Weighted Autoencoder Modeling (OWAM) framework. OWAM employs autoencoders for local outlier detection and generates correlation scores to assess neighboring traffic's influence. These scores serve as a weighted factor for neighboring sensors, before fusing them into the model. This information enhances the traffic model's performance and supports effective real-time updates, a crucial aspect for capturing dynamic traffic patterns. OWAM demonstrates a favorable trade-off between accuracy and efficiency, rendering it highly suitable for real-world applications. The research findings contribute significantly to the development of more efficient and adaptive traffic prediction models, advancing the field of transportation management for the future. The code and datasets of our framework is publicly available under \url{https://github.com/himanshudce/OWAM}.
\end{abstract}


\maketitle

\section{INTRODUCTION}

Traffic congestion is a well-known problem impacting public health, economic losses, and health problems \cite{su12114660}. It also contributes to environmental issues due to increased emissions and air pollution \cite{pasquale2015two}. As increasing road capacity may not always be feasible, optimizing traffic controllers becomes crucial for enhancing traffic flow. These systems can help alleviate congestion, reduce travel times, and enhance overall traffic flow.

Traffic modeling has advanced considerably, resulting in effective traffic prediction models in general. Yet, challenges emerge when these models encounter outlier traffic scenarios. In the realm of traffic, outliers refer to situations where data substantially deviate from the usual traffic patterns. These outliers can indicate abrupt traffic peaks, drops, or irregular trends, often arising from factors such as accidents, events, or roadwork. Furthermore, these traffic prediction models often lack generalizability for real-time predictions. Traffic is inherently dynamic, experiencing continuous fluctuations due to factors like accidents and special events. Emphasis should be placed on reducing training and evaluation time, ensuring the model's robustness for real-time updates and efficiency in handling real-world conditions.

\begin{figure*}[htbp]
    \centering
    \begin{subfigure}[b]{0.49\linewidth}
        \centering
        \includegraphics[width=\linewidth]{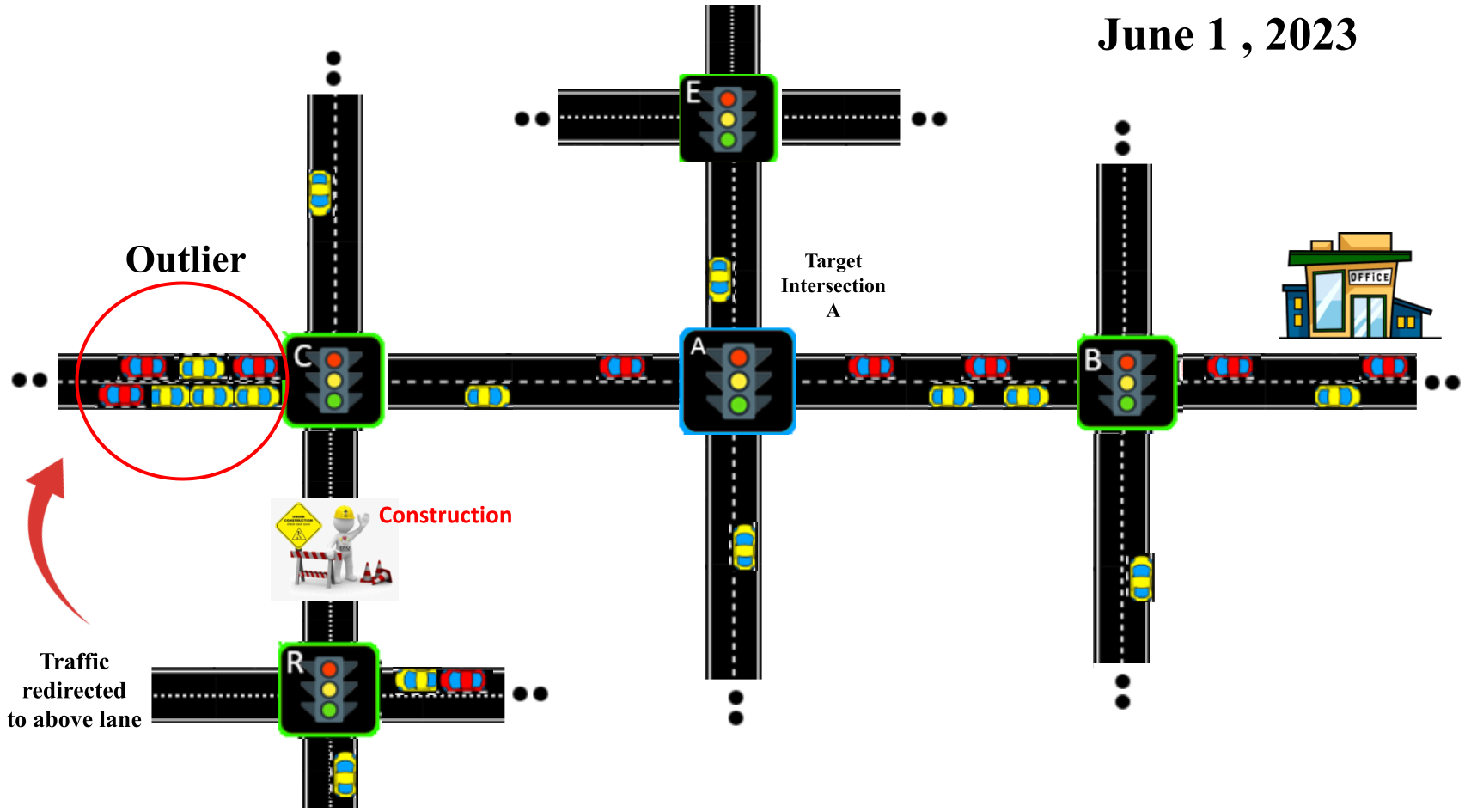}
        \caption{Traffic Condition on June 1 After an Accident}
        \label{fig:traffic_2}
    \end{subfigure}
    \hfill 
    \begin{subfigure}[b]{0.49\linewidth}
        \centering
        \includegraphics[width=\linewidth]{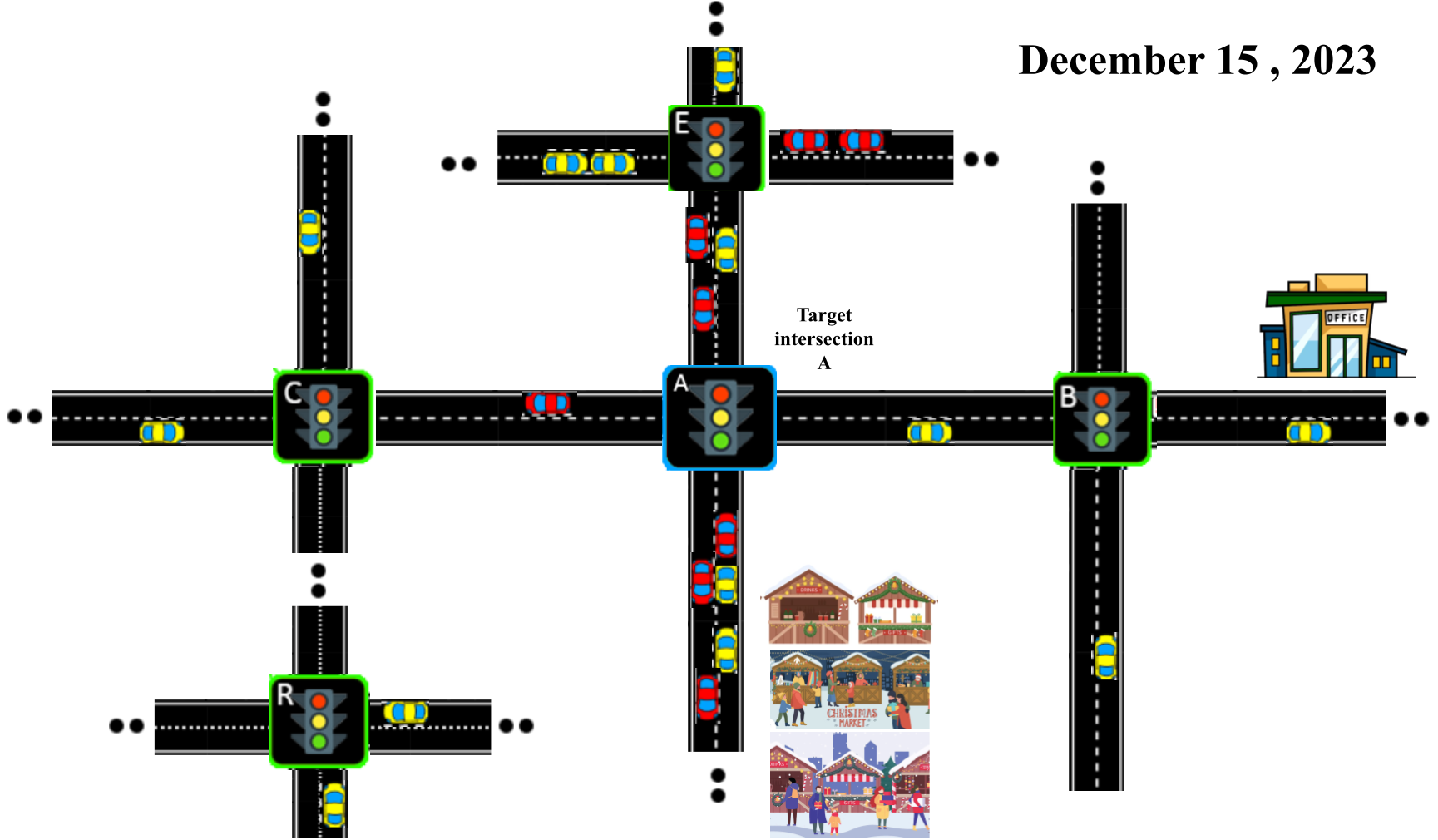}
        \caption{Traffic Condition on December 15, Around Christmas Time}
        \label{fig:traffic_3}
    \end{subfigure}
    \caption{Traffic Conditions of the City on Different Dates}
    \label{fig:traffic_conditions}
\end{figure*}

To illustrate these challenges, let's consider a city, where we aim to predict traffic conditions at Intersection A (Figure \ref{fig:traffic_conditions}), and, challenges arise due to abrupt changes in traffic patterns (Figure \ref{fig:traffic_2} \& \ref{fig:traffic_3}). The first case (Figure \ref{fig:traffic_2}) presents a clear outlier situation, while another case (Figure \ref{fig:traffic_3}) shows a total shift in traffic dynamics. Such instances can lead to long congestion at Intersection A if prediction models fail to accurately anticipate and identify future traffic flows. Current research falls short of offering a holistic solution that can simultaneously manage both outlier scenarios and real-time adjustments. This study seeks to bridge this gap by introducing the Outlier Weighted Autoencoder Modeling (OWAM) framework, a real-time model that aims to improve traffic predictions in both dynamic and outlier traffic scenarios simultaneously.

Building upon the groundwork laid by T. Mertens \cite{FPD-LOF}, our research basically aims to identify "relevant" intersections by assessing the influence of outliers from nearby sensors on the target sensor. We begin the process by leveraging Autoencoders (AE) under real-time settings to detect local outliers at each sensor. To enhance AE's performance further, we integrated Earth Mover's distance \cite{earth_mover}, a well-known distance metric, as a loss function in AE. These localized outlier scores then serve as the basis for calculating correlations between the target sensor and its neighboring counterparts. Subsequently, we merge the neighboring data with the target data, assigning weights based on their correlation scores to ensure precise influence. Furthermore, we incorporate outlier-based model updates through OWAM to effectively address both outlier scenarios and the dynamic changes simultaneously. To check the effectiveness of our proposed architecture we tested it on three real-world datasets, i.e. Hague (Section \ref{subsec:Hague}), METR-LA (Section \ref{subsec:Metrla}) and PEMS-BAY (Section \ref{subsec:PemsBay}).

The OWAM framework demonstrates superior performance compared to previous outlier-based approaches and the LSTM baseline. Though it falls short of surpassing the heavy graph-based techniques in terms of RMSE, it notably stands out in terms of runtime efficiency, rendering it highly suitable for real-world traffic management systems. This work extensively extends the presentation and the experimental study presented in \cite{OWAMSAC}. Overall, the following are the main contributions of our work:
\begin{enumerate}
    \item An innovative Outlier Weighted Autoencoder Modelling (OWAM) framework, that enhances traffic flow prediction in both outlier and real-time scenarios.
    
    \item A novel outlier detection architecture that integrates Earth mover's distance with Autoencoders to improve the accuracy of outlier identification.
    
    \item An extensive experimental evaluation over three real-world datasets to demonstrate the effectiveness of our proposed architecture on traffic prediction tasks utilizing traffic flow and traffic speed.
    
    \item Providing an open-source implementation of OWAM framework for reproducibility and further research.
\end{enumerate}

The paper is structured as follows: In Section \ref{sec:RELATED WORK}, we review relevant literature on outlier detection techniques and traffic modeling. Section \ref{sec:PAPF} provides essential background and mathematical formulation. Section \ref{sec:METHODOLOGY} presents the OWAM framework's design. Section \ref{sec:EXPERIMENTAL SETUP} describes datasets and metrics. Section \ref{sec:RESULTS} discuss our findings on traffic prediction. Section \ref{sec:CONCLUSION AND FUTURE WORK} summarizes the discussion and outlines future research directions.

\section{RELATED WORK}
\label{sec:RELATED WORK}

This Section explores relevant literature, focusing on outlier detection techniques and traffic modeling, both of which are central to our research objective. Outlier detection involves identifying data points that significantly deviate from the norm, with various techniques applied across domains. These methods include One-Class Support Vector Machines (OC-SVM) \cite{OC-SVM}, Local Outlier Factors (LOF) \cite{LOF} and Isolation Forests \cite{IF}. However, most of these approaches are not well-suited for real-time applications like traffic flow forecasting due to high response times, scalability issues, and difficulties in handling high-dimensional real-time data. To handle such dynamic tasks, several contributions have been made in the realm of online anomaly detection for data streams. Notable approaches include Half-Space Trees (HST) \cite{HST}, Robust Random Cut Forest (RRCF) algorithm \cite{RRCF} and the extension of the Local Outlier Factor (LOF) method for data streams \cite{ILOF}.

Nevertheless, Autoencoders (AEs) have proven to be the most effective method in both offline and online scenarios. The KitNET algorithm \cite{Kitnet} efficiently identifies network intrusions through an ensemble of autoencoders. MemStream \cite{Memstream} introduces a novel approach by combining denoising autoencoders with a memory module to address concept drift and memory poisoning in real-time outlier detection for streaming data. Additionally, to tackle concentrated bursts and a higher proportion of anomalies in data streams, Probability Weighted Autoencoders (PW-AE) \cite{PWAE} are proposed, where the loss function is optimized based on the probability of an instance being an anomaly. Not only does it handle evolving data streams and concept drift, but it also outperforms previous models in terms of accuracy and efficiency. In this paper, we build upon this promising prior work \cite{PWAE} and further modify it for enhanced outlier detection, as detailed in subsequent sections.

In the domain of traffic modeling, various approaches have been explored to enhance prediction accuracy and efficiency. Broadly categorized, traffic flow prediction models fall into three main groups: parametric techniques, machine learning techniques, and deep learning techniques \cite{RW_Overview}. Parametric techniques use statistical methods with parameters to model data like Kalman filters \cite{emami2020shortterm} and ARIMA models \cite{box1970time} for time series analysis. However, these methods may not capture complex data patterns as effectively as machine learning or deep learning techniques. Machine learning techniques, such as Support Vector Machines (SVM) \cite{scale_lof} and K-nearest neighbors (KNN) \cite{knn}, consider spatial characteristics for congestion prediction. Some incorporate dynamic distance measures and state matrices for accuracy improvement. Deep learning employs neural networks with multiple layers to learn complex data patterns. They extract space-time dependencies from heterogeneous data sources, enhancing accuracy. For example, Convolution Neural Networks (CNNs) extract spatial features from traffic images, while LSTMs capture temporal dependencies in traffic flow data. Encoders reduce data dimensionality and Graph Convolution Networks (GCNs) model relationships between regions. Moreover, hybrid models, combining multiple techniques also boost performance \cite{li2018diffusion, kouziokas2021deep}. 

Outlier-based hybrid approaches offer valuable insights into traffic modeling by focusing on outlier detection in traffic flow data. Polson et al. \cite{POLSON20171} demonstrated the effectiveness of neural networks in handling outlier scenarios in traffic data. On the other hand, LOF-based models have been proposed to detect outliers within Probability Distributions of traffic flows (FPDs) \cite{LOF_OUT, FPD, COMPSAC2021}. Previous studies have also investigated different FPD-based distance metrics for effective outlier detection, with Earth Mover's distance \cite{earth_mover} proving to be the most effective. 

Prior outlier-based approaches have been constrained by traditional outlier detection methods, predominantly tested on similar datasets, and have been limited in their ability to intelligently integrate information. Additionally, researchers largely overlooked the real-time aspect of traffic,  where traffic consistently evolves and adapts patterns in response to various ongoing events, a critical aspect demanding further investigation. Regular model updates are vital to ensure accurate and up-to-date traffic models that reflect real-time changes in traffic flow. These models should strike a balance between accuracy and efficiency to facilitate real-world scenarios.

\section{PRELIMINARIES AND PROBLEM FORMULATION}
\label{sec:PAPF}

Traffic controllers ($TC$) manage traffic signal timing at intersections, optimizing various traffic movements. They rely on data from multiple traffic monitoring and management ($TM$) systems, which collect real-time traffic information at each sensor $I$, including vehicle volume, speed, and occupancy. Our objective is to enhance the prediction accuracy of the traffic management systems $TM_{I}$, assisting $TCs$ in decision-making and optimizing traffic flow. We trained a model $M$ with both offline and online objectives for comparison and real-time applicability assessment. 

Let $L^+ = \{(x(i), y(i)) \,|\, \forall i \in S\}$ be a sequence of $S$ tuples, where each tuple consists of a feature-vector $x(i) \in {R}^d$ and a corresponding real value to be predicted $y(i) \in R$. Additionally, let $L_{\text{train}} = \{(x(i), y(i)) \,|\, \forall i \in \{1, \ldots, P\}\}$ be a sub-sequence used for training the model $M$. Let $L_{\text{test}} = \{(x(i), y(i)) \,|\, \forall i \in \{1, \ldots, m\}\}$ be the sub-sequence used for testing the model.

Here, $L_{\text{train}}$ and $L_{\text{test}}$ are disjoint subsets of $L^+$. Given a function $f: {R}^d \rightarrow {R}$ that was trained on $L_{\text{train}}$ with P samples, the objective can be formulated with $min(z_1)$ such that -
\begin{equation}
z_1 = \frac{1}{N} \sum_{i=1}^{N} \left( \mathcal{L}_r, L_{\text{test}}, f(L_{\text{train}}) \right)
\end{equation}
Where N denotes the total number of sensors and  $\mathcal{L}_r$ represents a regression loss.

In addition to the regular (offline) settings, we incorporated a real-time incremental updating strategy \cite{incremetal} with time windows $T$ of different lengths. Let $L_{train}$ contain the data up to time point $t$, and $L_{test} = \{(x(i), y(i)) \,|\, \forall i \in \{t+1, \ldots, t_{n}\}\}$ represent infinite sub-sequences with interval length $T$. For a given validation protocol $\mathcal{V}$, the problem can be formulated by $min(z_2)$ such that -
\begin{equation}
z_2 = \frac{1}{N} \sum_{i=1}^{N} \sum_{j=1}^{t_n} \mathcal{V}\left( \mathcal{L}_r, L_{\text{test}}, f(L_{\text{train}}) \right)
\end{equation}

For the validation protocol $\mathcal{V}$, we employ the test-then-train \cite{bifet2018machine} approach. In this evaluation method, models are first validated based on a performance metric and subsequently trained on each instance within the time window ${T}$ of $L_{test}$ at its arrival time.

To achieve this objective, we introduce the OWAM (Outlier Weighted Autoencoder Model) framework. We outline its key components in details below for through understanding.

\paragraph{\textbf{Flow Probability Distribution (FPD)}}

Flow Probability Distributions (FPDs) \cite{FPD-LOF} represent the probability of traffic occurrences within specific time intervals, derived from real traffic data. FPDs encapsulate traffic distribution patterns and can be created for vehicle flow, average speed or any relevant traffic data, to get a comprehensive understanding of traffic dynamics.

Let us denote the FPD for a specific sensor B during the time period H as $FPD(HB)$. To calculate $FPD(HB)$, we start with a collection of aggregated traffic flow values, denoted as $X_{HB}$:
\begin{equation}
X_{HB} = (x_{hB1}, x_{hB2}, \ldots, x_{hBH})
\end{equation}

Here, $x_{hB}$ represents the aggregated traffic flow or speed value for the intersection B during the $h$-th time interval, with $h$ ranging from 1 to $H$.

To calculate the FPDs, we apply the formula as created by Zimek and Djenouri in \cite{LOF}:
\begin{equation}
\label{eq:fpds_equation}
FPD(HB) = \frac{\sum_{x_h \in X_H} P(x_h)}{P(x_{hB})}, \quad \forall x_{hB} \in X_{HB}, \quad h = 1, \ldots, H
\end{equation}

\paragraph{\textbf{Outlier Detection using Autoencoders}}

This research explores various outlier detection techniques, including well-established methods like Half-Space-Tree (HST) and Kit-net (see Section \ref{sec:RELATED WORK}). However, our primary focus is on the Autoencoders, recognized for their superior online anomaly detection capabilities.

\begin{figure}[htbp]
    \centering
    \includegraphics[width=0.8\columnwidth]{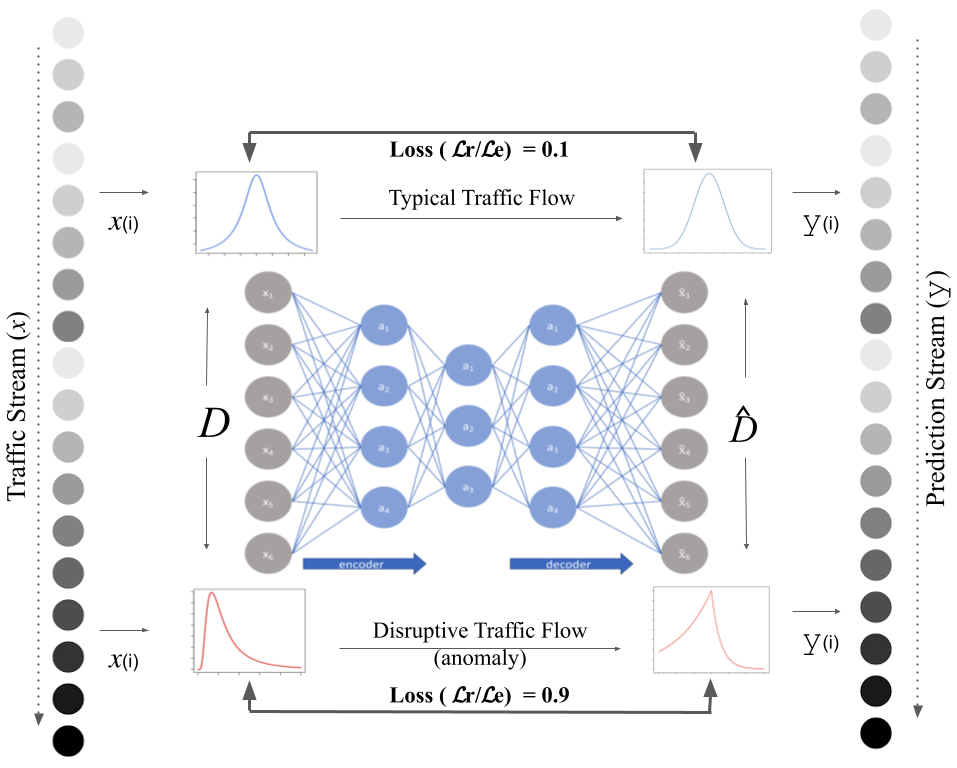}
    \caption{Autoencoders (Online): Outlier Detection}
    \label{fig:AE}
\end{figure}

Autoencoders, commonly used for anomaly detection, aim to reconstruct $\hat{D}$ input data $D$ by encoding it into a lower-dimensional representation and then decoding it back to its original format (Equation \ref{eq:AEeq}). Anomalies, deviating significantly from the training data, result in higher reconstruction errors/loss (Equation \ref{eq:LossAE}). As shown in Figure \ref{fig:AE}, our research includes training AE on continuous traffic stream of FPDs in real-time and if faced with an anomaly, say, characterized by a left-skewed distribution with substantial magnitude (Figure \ref{fig:AE}), the model generates a higher reconstruction loss, indicating the presence of an outlier. 
\begin{equation}
\label{eq:AEeq}
\hat{D} = Decoder(Encoder(D))
\end{equation}
\begin{equation}
\label{eq:LossAE}
\textit{Reconstruction Loss}(\mathcal{L}) = \mathcal{L}_{r/e}(\hat{D}, {D})
\end{equation}

\begin{figure*}[htbp]
    \centering
    \includegraphics[width=0.8\textwidth]{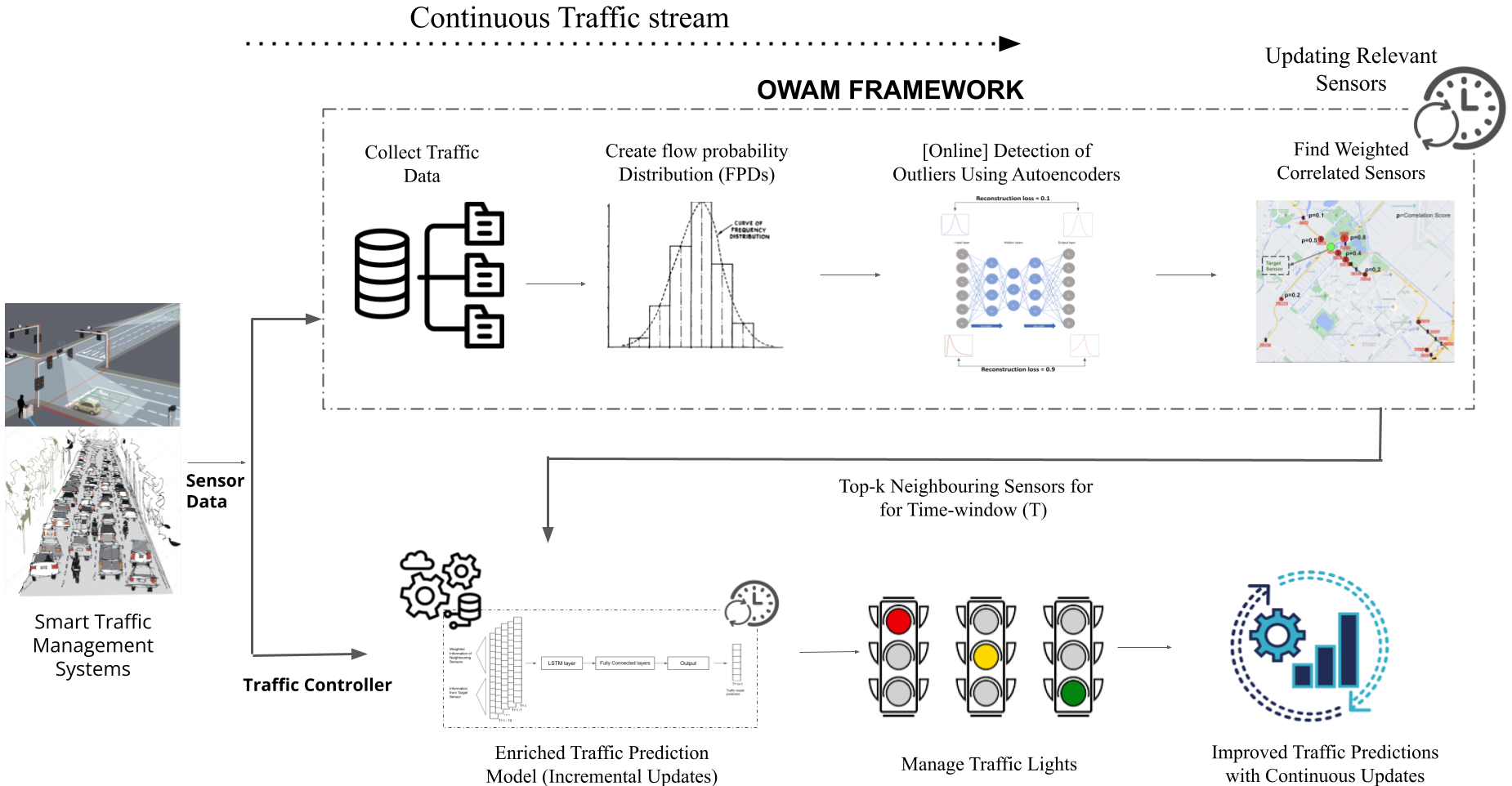}
    \caption{Overview of OWAM Framework with Real-Time Updates}
    \label{fig:owri_dynamic}
\end{figure*}

To train the autoencoder, we aim to minimize the reconstruction loss ($\mathcal{L}_r$), typically measured using Root Mean Square Error (RMSE). Building upon the previous work by Bazan et al. \cite{bazan2019quantitative}, where they found that Earth Mover's distance (EMDs) exhibited superior performance in comparing two distributions, we utilized this knowledge and integrated EMD as a loss function $\mathcal{L}_e$ in Autoencoders as can be depicted in the Figure \ref{fig:AE}, where loss can be used either as RMSE ($\mathcal{L}_r$) or EMD ($\mathcal{L}_e$).  

Earth's Mover Distance (EMD) \cite{earth_mover} is well-suited for capturing complex differences between entire distributions, in contrast to Euclidean distance (RMSE), which emphasizes point-wise errors. EMD quantifies the minimum effort required to transform one distribution into another, considering their overall structure, and is less influenced by individual extreme points.

Finally, this research incorporates the Long Short-Term Memory (LSTM) model \cite{LSTM}, a widely used sequence prediction model, in its existing form. It is important to highlight that our proposed framework can be integrated with any traffic modeling framework, as we do not alter the original model but enhance its performance by passing only crucial information.

\section{METHODOLOGY}
\label{sec:METHODOLOGY}

This section presents a comprehensive overview of our novel framework, Outlier Weightage Autoencoder Modeling (OWAM), along with an elaboration on the rationale behind its design choices. 

Figure \ref{fig:owri_dynamic} outlines our comprehensive traffic modeling approach. Initially, traffic data is collected at 5-minute intervals. Subsequently, Flow Probability Distributions (FPDs) are generated from this data, and these distributions undergo online AE-based outlier detection to compute local anomaly scores for each sensor. A Pearson's correlation coefficient is then employed to calculate the correlation score between the target sensor and its neighboring sensors to determine their influence. These correlation scores are used as weights to adjust the importance of the neighboring sensor information before integrating it with the data from the target sensor. This integrated information is subsequently fed into an LSTM model for efficient traffic prediction. It is important to note that not only LSTM but any other modeling framework can benefit from the weighted information generated by OWAM. To address real-time changes, we continually recalculate outlier information using online AE for every $T$ time window and incrementally update the LSTM model with this revised information.  Each component is discussed below to provide a comprehensive understanding of the OWAM framework.

\subsubsection{Creating Flow Probability Distribution (FPDs)}
The first step involves creating Flow Probability Distributions (FPDs) from the collected data (Equation \ref{eq:fpds_equation}), which offers several advantages. FPDs help smooth out short-term fluctuations and noise, simplifying analysis, and capturing higher-level traffic patterns and trends. They enhance the identification of peak hours, daily variations, and weekly trends, providing a better understanding of traffic dynamics. Additionally, FPDs are more robust to outliers compared to individual 5-minute intervals, improving analysis reliability. Choosing the right time window for FPDs is crucial, balancing pattern capture and noise reduction. Very short intervals can introduce excessive noise, while very long intervals may over smooth data, potentially missing patterns.

\begin{figure}[htbp]
    \centering
    
    \begin{subfigure}[b]{0.30\linewidth}
        \includegraphics[width=\linewidth]{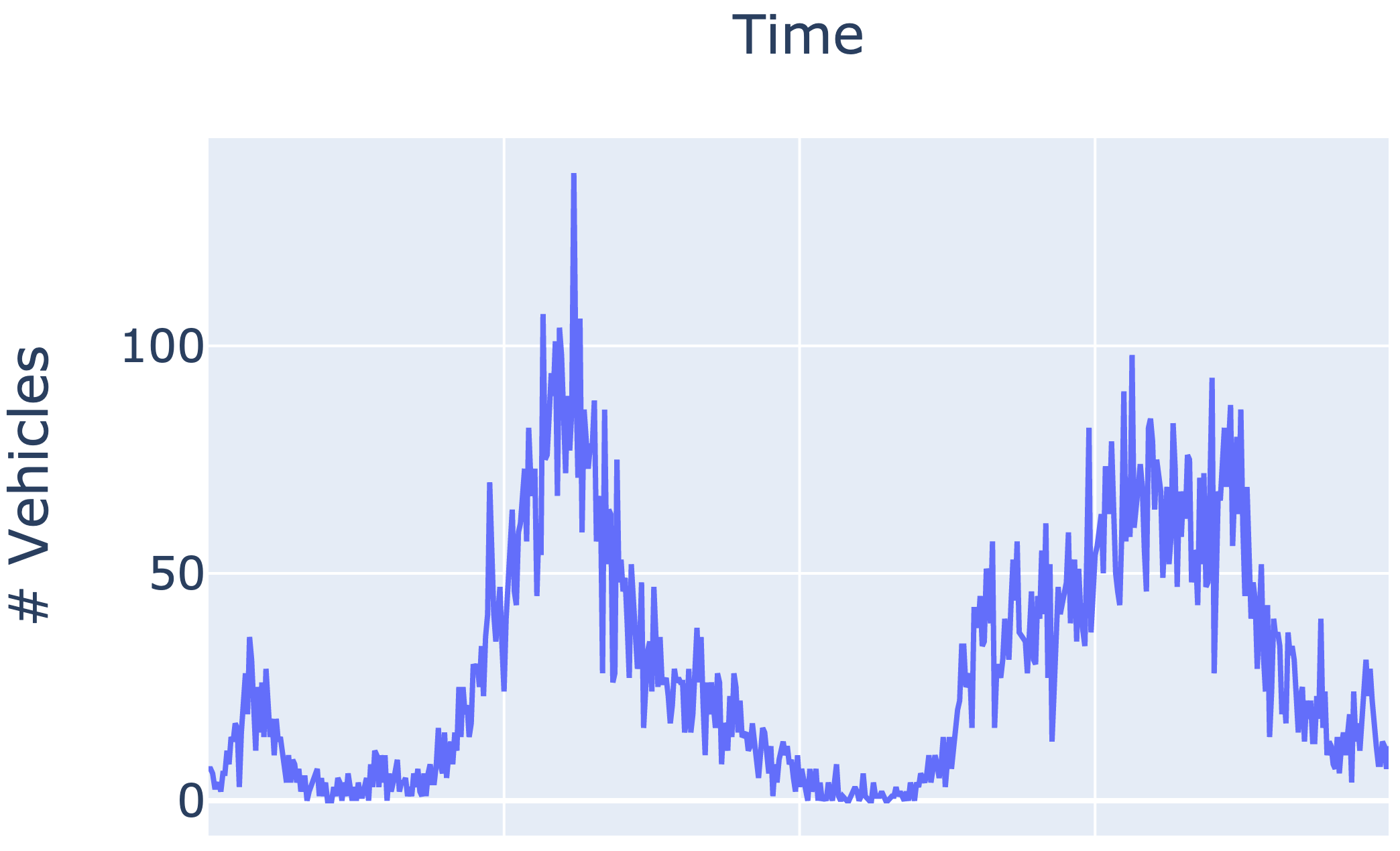}
        \caption{5-Minute Window}
        \label{fig:hague-5min}
    \end{subfigure}
    \hfill
    \begin{subfigure}[b]{0.30\linewidth}
        \includegraphics[width=\linewidth]{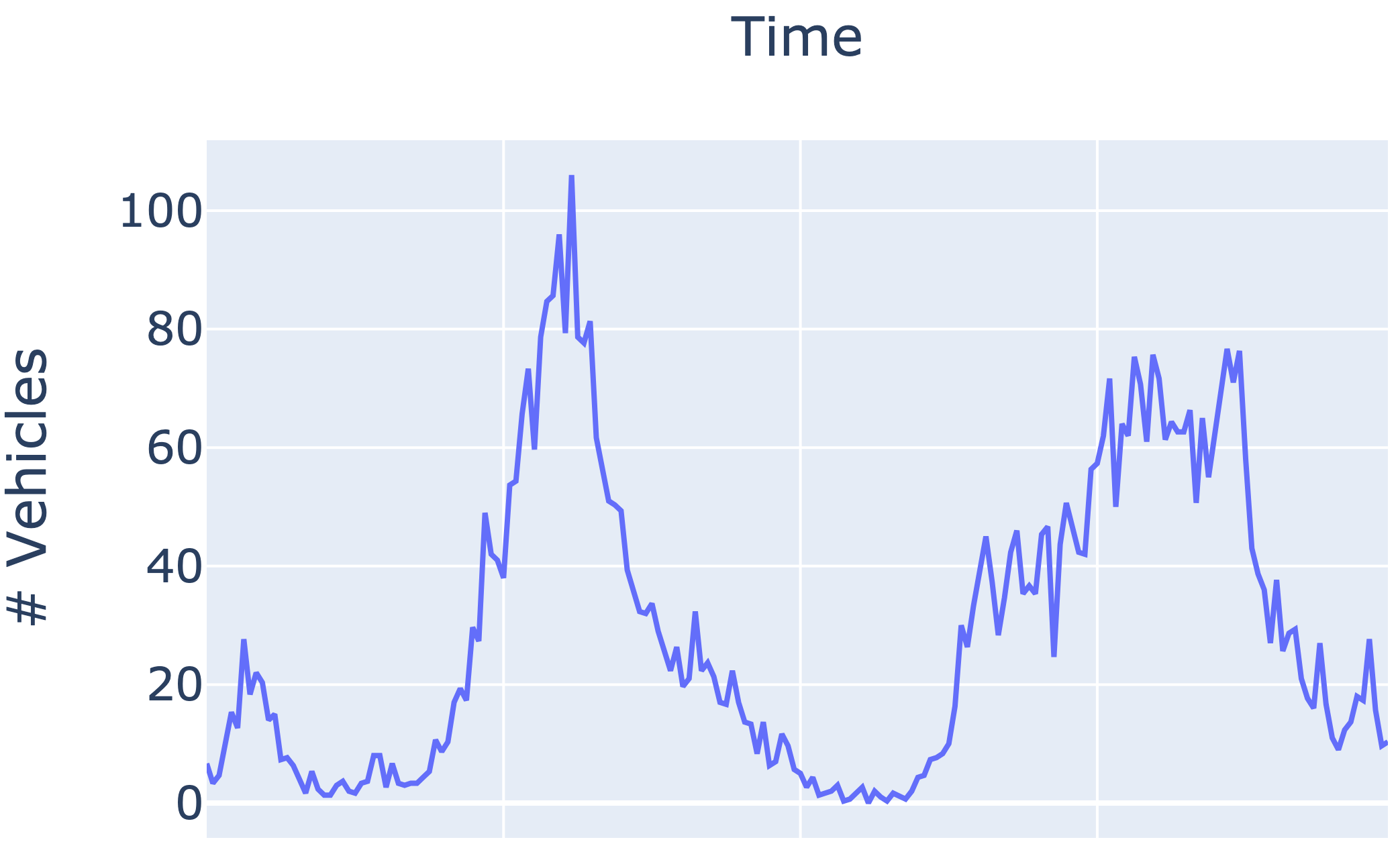}
        \caption{15-Minute Window}
        \label{fig:hague-15min}
    \end{subfigure}
    \hfill
    \begin{subfigure}[b]{0.30\linewidth}
        \includegraphics[width=\linewidth]{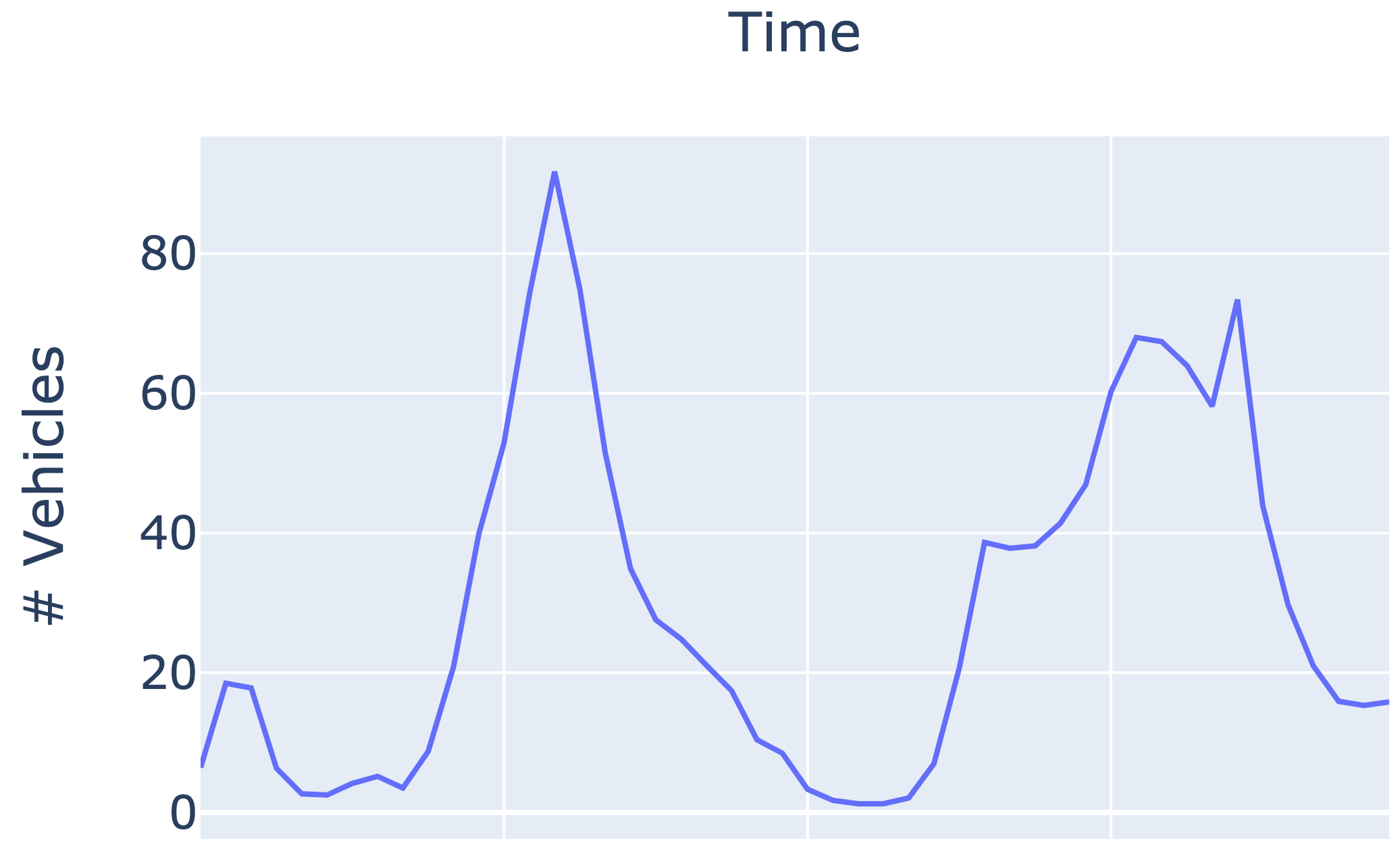}
        \caption{60-Minute Window}
        \label{fig:hague-1H}
    \end{subfigure}
    
    \caption{Aggregation Plot at 5, 15, and 60 Minutes - Sensor K504}
    \label{fig:hague-vehicle-counts}
\end{figure}

Figure \ref{fig:hague-vehicle-counts} illustrates these interpretations. Figure \ref{fig:hague-5min} displays data aggregated every 5 minutes with noticeable noise and fluctuations. Although a 15-minute interval aggregation (Figure \ref{fig:hague-15min}) reduces some noise, it still remains. In contrast, the 1-hour window (Figure \ref{fig:hague-1H}) makes it easier to identify outliers and patterns. So we aggregated data into 1-hour FPDs using 5-minute intervals, for optimizing outlier detection.

\subsubsection{Autoencoders for Outlier Detection}
\label{subsec:AEWORK}
In the second step, the continuous stream of Flow Probability Distributions (FPDs) undergoes online anomaly detection to identify outliers at each sensor. Utilizing online autoencoders, which adapt to incoming data in real-time, allows continuous updates of weights and parameters to capture evolving patterns and anomalies without the need for retraining on the entire dataset. Probability-weighted autoencoders (PW-AE) were primarily employed for their superior performance, as detailed in Section \ref{sec:RELATED WORK}. Additionally, we introduce a novel PW-AE-EMD model, integrating Earth Mover's distance (EMD) loss ($\mathcal{L}_e$) to measure dissimilarity between the input FPD ($D$) and the reconstructed FPD ($\hat{D}$) for enhanced outlier detection (Equation \ref{eq:AEeq} \& \ref{eq:LossAE}).

We preserve the outlier scores for each sensor without applying additional threshold filtering. The use of filtering can alter the correlation strength by discarding valuable information about how neighboring sensors were influenced by outliers and their subsequent return to normal behavior. By maintaining the raw outlier scores, we gain a comprehensive understanding of the dynamics of outliers and their effects on nearby sensors. This approach enhances our analysis of traffic patterns and behavior.

\subsubsection{Weighted Correlation Analysis for Relevant Sensor Identification}

To proceed, it is essential to have target sensors for correlation analysis and prediction. In a real-world scenario, we would need to predict each sensor individually, making every sensor a potential target. However, due to computational constraints, we opted for a more manageable approach by selecting representative samples from the data.  

For the Hague dataset (refer to Section \ref{subsec:Hague}), we employed the same target sensors as used in a previous outlier-based study \cite{FPD-LOF} due to their strategic positioning on the lane. However, in the case of the remaining two datasets i.e. METR-LA (Section \ref{subsec:Metrla}), and PEMS-BAY (Section \ref{subsec:PemsBay}), the exact sensor locations were unavailable, making it impossible to apply this domain-specific knowledge. Consequently, in these cases, we randomly selected five target sensors from each dataset to act as representatives for our study.

For each target sensor, we compute correlation scores with other sensors using their outlier scores. Pearson's correlation coefficient serves as the metric for this purpose. While, in an ideal scenario, we could consider introducing a time lag to account for traffic propagation, but the datasets used in our study contained closely situated sensors. Consequently, traffic could traverse between them within relatively short time intervals. In the Hague dataset, for instance, the average travel time between any two sensors was consistently less than 1 hour. Although we lacked precise travel time information for the PEMS-BAY and METR-LA datasets, available distance data suggested their relative proximity. In fact, the maximum distance between the farthest sensors in these datasets was approximately 20 km.

Utilizing correlation scores, we introduced the concept of outlier-weighted correlation integration. Instead of directly incorporating information from correlated neighboring sensors, we assign weights based on their correlation coefficients. These weights quantify the influence of each neighboring sensor on the target sensor. In Figure \ref{fig:owam_corr}, the green circle represents the target sensor, while the neighboring sensors are depicted as red circles with varying intensities corresponding to their correlation scores. When transmitting information from these sensors, along with the target sensor's data, to the LSTM model, we apply a weighted approach. Specifically, we multiply the transmitted information by weights determined by their respective correlation scores. This weighting mechanism allows the model to emphasize more influential sensors based on their outlier characteristics, resulting in enhanced traffic predictions.

\subsubsection{Traffic Prediction}

\begin{figure}[htbp]
    \centering
    \begin{subfigure}[b]{0.7\linewidth}
        \centering
        \includegraphics[width=\linewidth]{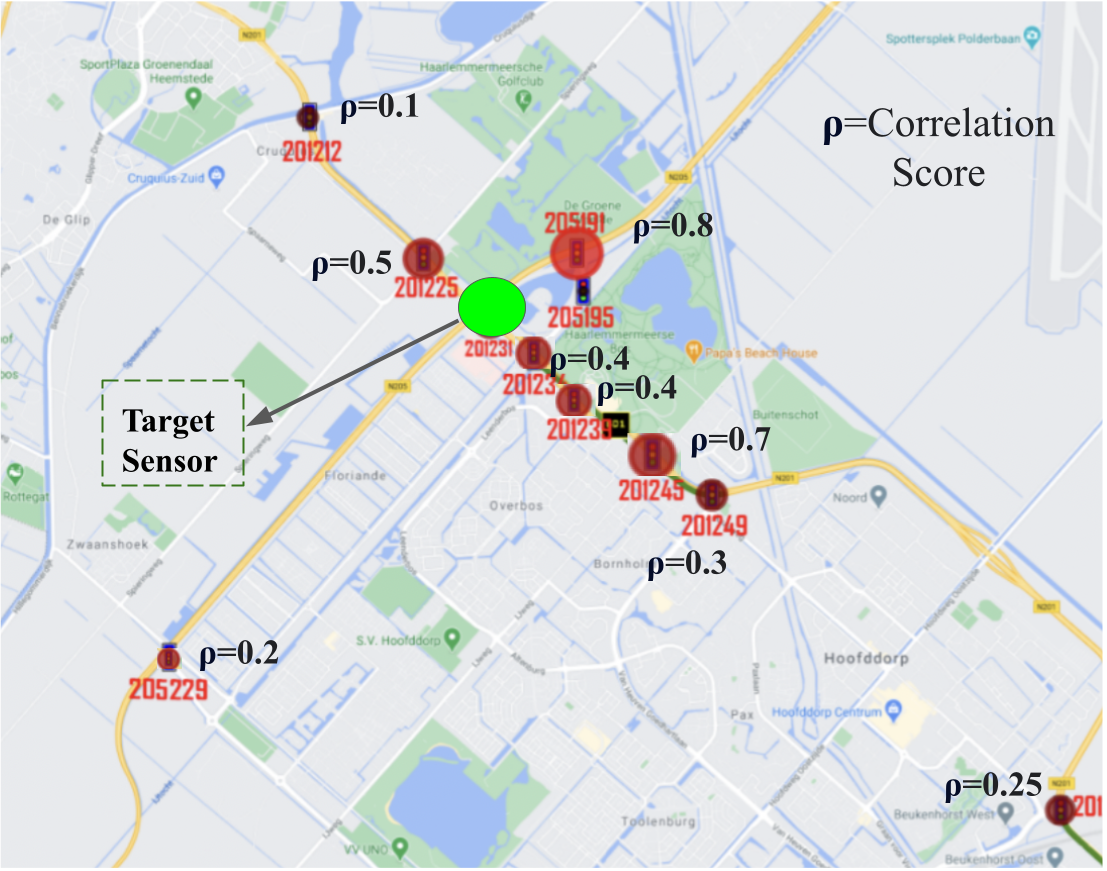}
        \caption{Correlated Weighted Sensors}
        \label{fig:owam_corr}
    \end{subfigure}

    \begin{subfigure}[b]{0.9\linewidth}
        \centering
        \includegraphics[width=\linewidth]{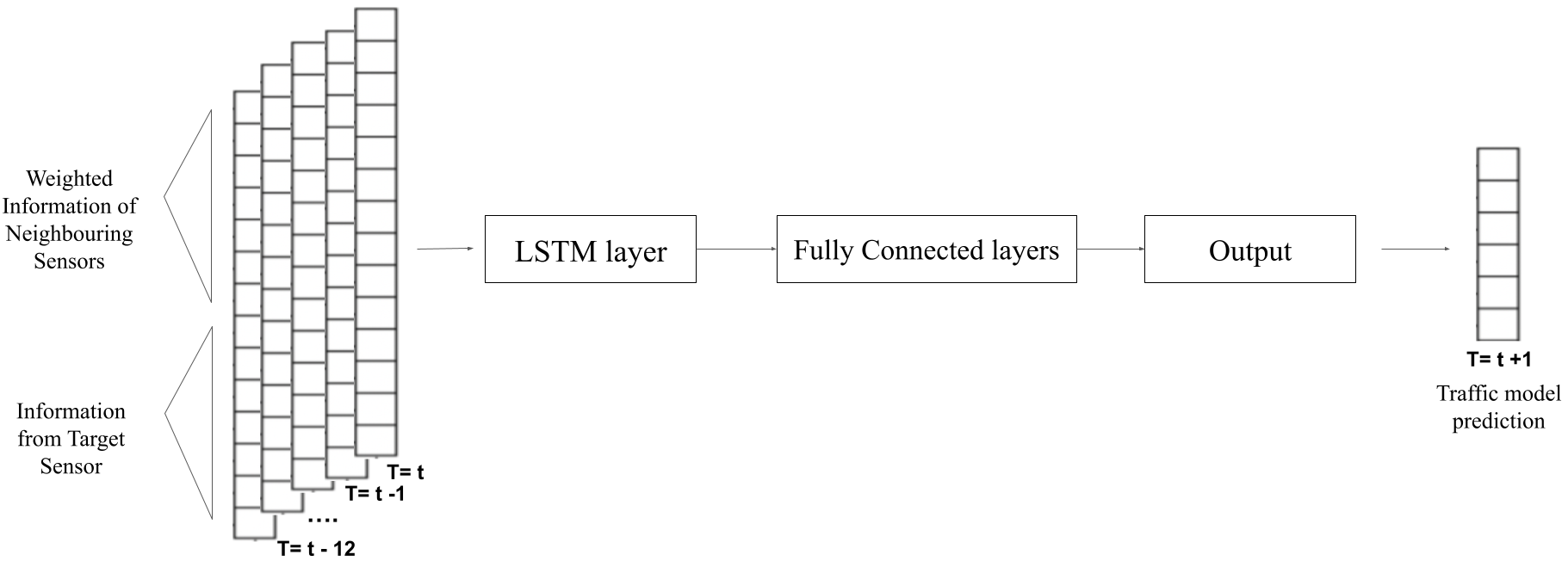}
        \caption{LSTM Overview for Traffic Prediction}
        \label{fig:lstm}
    \end{subfigure}
    \caption{Correlated Weighted Sensor and LSTM Overview}
    \label{fig:owam_combined}
\end{figure}

After obtaining the weighted correlation information from neighboring sensors, we concatenated their data with the target sensor's data and inputted it into the LSTM model, as illustrated in Figure \ref{fig:lstm}, for predicting the traffic flow of the target sensor. As mentioned earlier, this weighted information from neighboring sensors can be utilized by any traffic prediction model, not exclusive to LSTM. To explore the impact of dimensionality, our experiments involved various sensor subsets, ranging from selecting only the top $k$ sensors to utilizing all neighboring sensors. The threshold ($\theta$) determined the selection of the top $k$\% of sensors. For instance, choosing the top 5\% ($\theta=0.05$) in the METR-LA dataset with 207 sensors implied selecting the top 10 neighboring sensors for each target sensor.

In offline settings, we partitioned the data into an 80/20 ratio for model training and evaluation. The LSTM model processed the data sequentially, with input vectors ($x_{i}$) representing the previous 1-hour window (12 timestamps, 5-minute intervals), and output vectors ($y_{i}$) representing the subsequent 5-minute traffic situation (Speed/Number of vehicles) to be predicted. We conducted the comparative analysis in offline settings for fair comparison purposes, primarily because there were no online versions available for the other models being evaluated.

\subsubsection{Real-Time Traffic Modelling}

To accommodate the dynamic nature of traffic, we have implemented an incremental updating strategy, as depicted in Figure \ref{fig:owri_dynamic}. Correlated sensors are periodically refreshed at specific intervals referred to as time windows ($T$), and this updated information is utilized for incremental updates to the LSTM model. Initially, the base model is trained using the first 50\% of the data, with the target sensor having access to this historical data up to time point $t$. Subsequently, we incrementally update the model with the next $t_n$ time windows of interval $T$. For each window at $t+1$, we obtain the outlier scores using online AE (Secition \ref{subsec:AEWORK}), recalculate the correlation score (weights), and adjust the significance of neighboring sensors with incoming data. Prior to updating the LSTM model, we evaluate its performance and then train it with the data from the current window, following the test-then-train methodology \cite{bifet2018machine}.

We experimented with various time windows, ranging from 1 hour to 1 month, to assess both short-term and long-term impacts. We compare three scenarios: outlier-weighted updates (incremental updates using OWAM), standard real-time updates (incremental updates without OWAM, keeping neighboring sensors fixed), and an offline model with no updates post-training. Through this comparison, our goal is to demonstrate the superiority of real-time updates and underscore the advantages of incorporating outlier-weighted correlations in traffic prediction.

\section{EXPERIMENTAL SETUP}
\label{sec:EXPERIMENTAL SETUP}

This section presents the following key components: performance metrics, an overview of the datasets along with their statistics, and system details. These elements serve as the foundation for conducting our experiments.

\subsection{Performance Metric}
\subsubsection{RMSE Metric}
\label{sec:RMSE}

We employed the Root Mean Squared Error (RMSE) metric, a common measure for regression problem evaluation. RMSE quantifies the average difference between predicted values ($\hat{y}_i$) and true values ($y_i$) in a dataset. It is computed as: $RMSE = \sqrt{\frac{1}{n}\sum_{i=1}^{n}(\hat{y}_i - y_i)^2}$.


\subsubsection{Temporal Analysis} 
In addition to score metrics, we investigated model convergence and prediction speed during training and testing. Understanding how quickly models learn and make predictions is vital for real-time applications, where timely decisions are essential. Assessing convergence time also gauges efficiency and effectiveness in learning from data.

\subsection{Dataset}
\label{sec:data}
This research utilizes three datasets to investigate traffic patterns: The Hague dataset, which provides information on the aggregated number of vehicles every 5 minutes, and two widely referenced datasets in the literature, METR-LA, and PEMS-BAY, which capture the average speed of traffic in 5-minute intervals. All three datasets considered as traffic flow measurements in this study.

\subsubsection{Hague Dataset}
\label{subsec:Hague}
The Hague dataset, obtained from 23 intersections in The Hague, Netherlands, spans from January 1, 2018, to March 31, 2020, with 5-minute intervals (Table \ref{tab:dataset-info}). The dataset had minimal missing values (less than 0.01\%), which were imputed with linear interpolation. Sensor distances and average travel times were obtained from the Google Maps API and used as input for Graph Neural Networks \cite{DGCRN}.

\subsubsection{METR-LA Dataset}
\label{subsec:Metrla}
The METR-LA dataset is a widely used benchmark dataset for traffic modeling and forecasting. The data is gathered from the highway of Los Angeles containing 207 selected sensors indicating speed in miles per hour and ranging from March 1st, 2012 to June 30th, 2012 as indicated in Table \ref{tab:dataset-info}. It covers multiple sensor locations across different highways, providing a diverse and representative set of traffic patterns for analysis. 

\begin{figure*}[htbp]
    \centering
    \caption{Distributions of Speed, Number of Vehicles, and Inter-node Correlations}
    \begin{subfigure}[b]{0.30\textwidth}
        \centering
        \includegraphics[width=\textwidth]{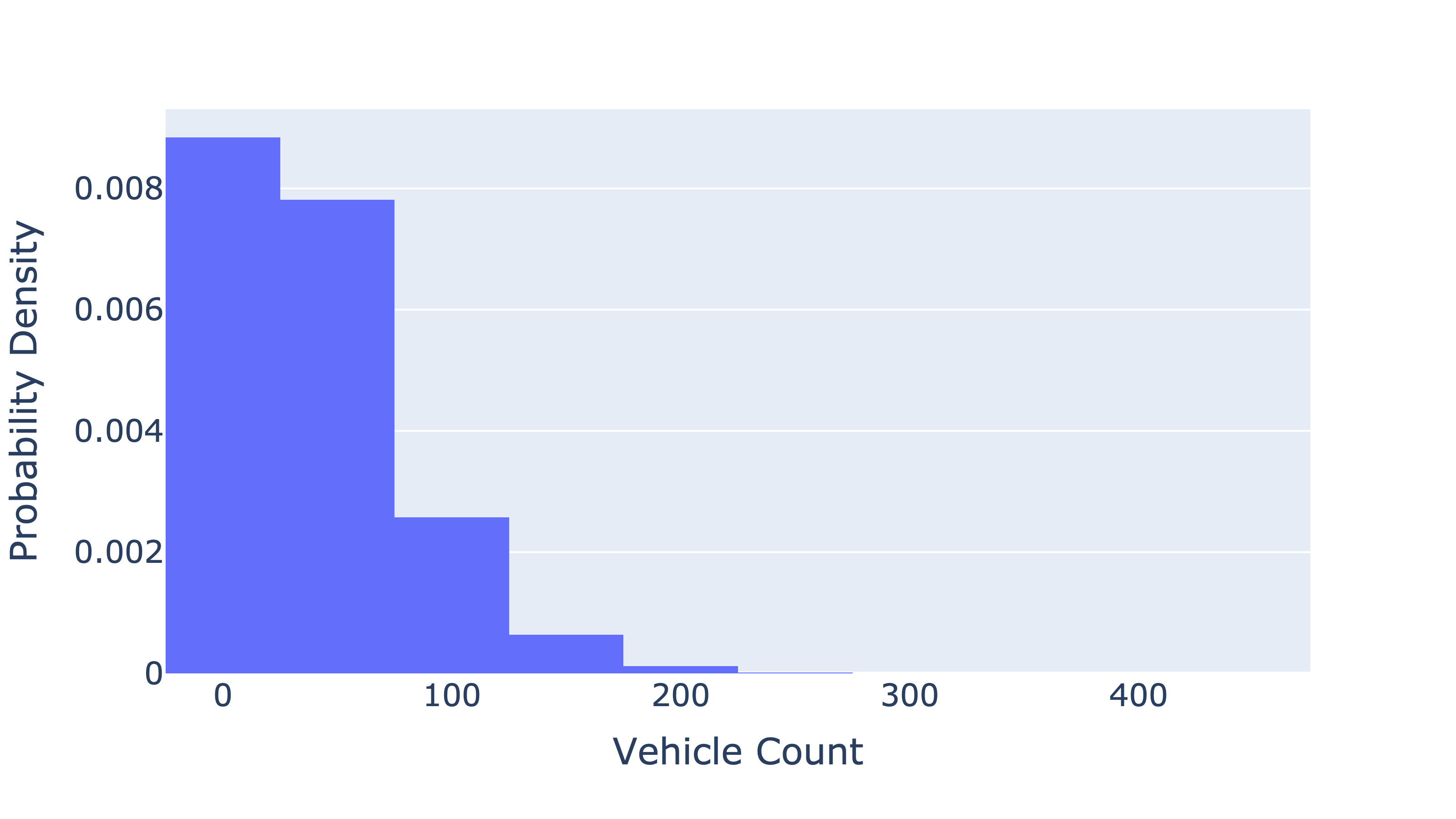}
        \caption{Hague \#Veichle Distribution}
        \label{fig:hague-number}
    \end{subfigure}%
    \hfill
    \begin{subfigure}[b]{0.30\textwidth}
        \centering
        \includegraphics[width=\textwidth]{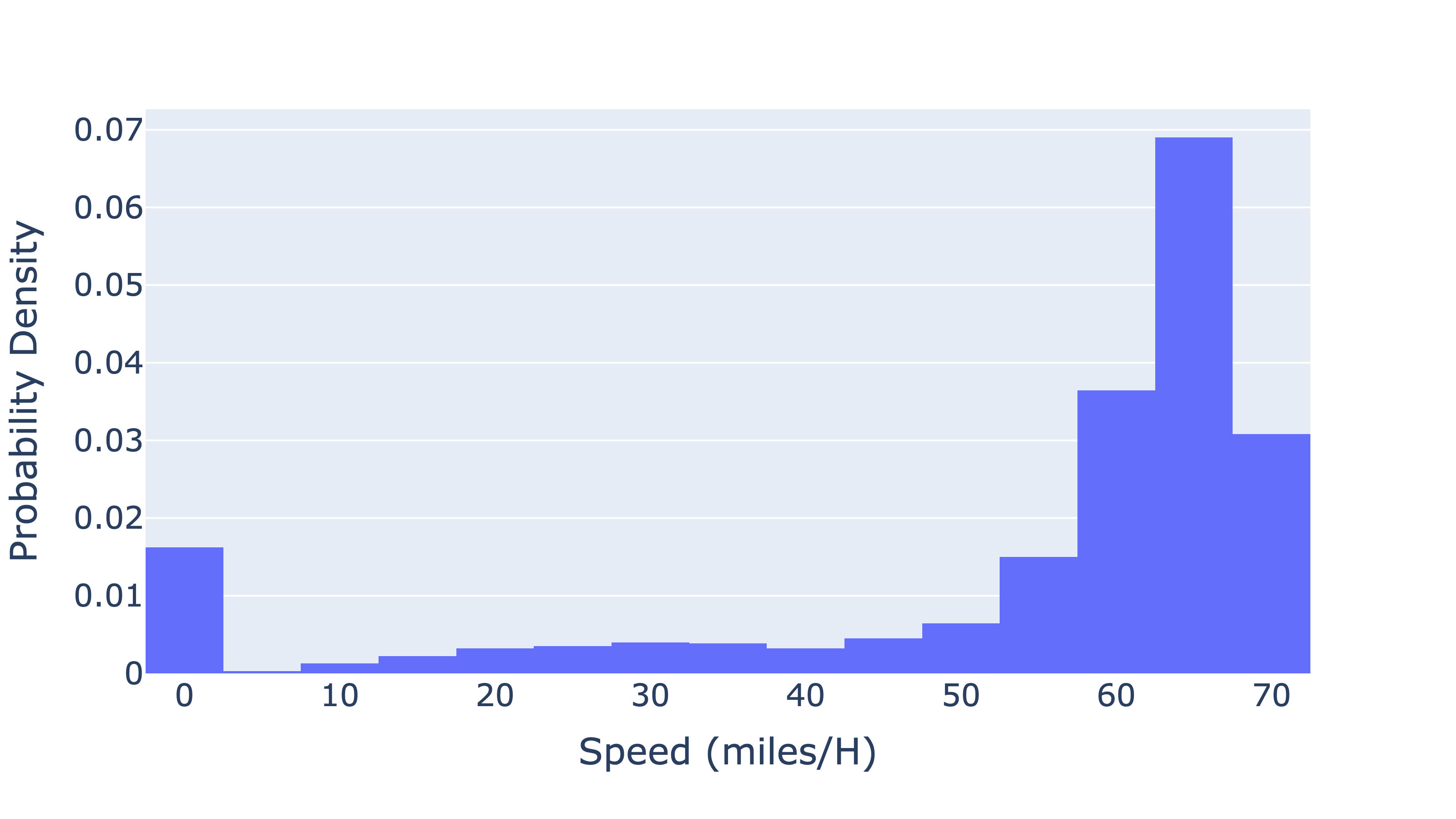}
        \caption{METR-LA Speed Distribution}
        \label{fig:metr-la-speed}
    \end{subfigure}%
    \hfill
    \begin{subfigure}[b]{0.30\textwidth}
        \centering
        \includegraphics[width=\textwidth]{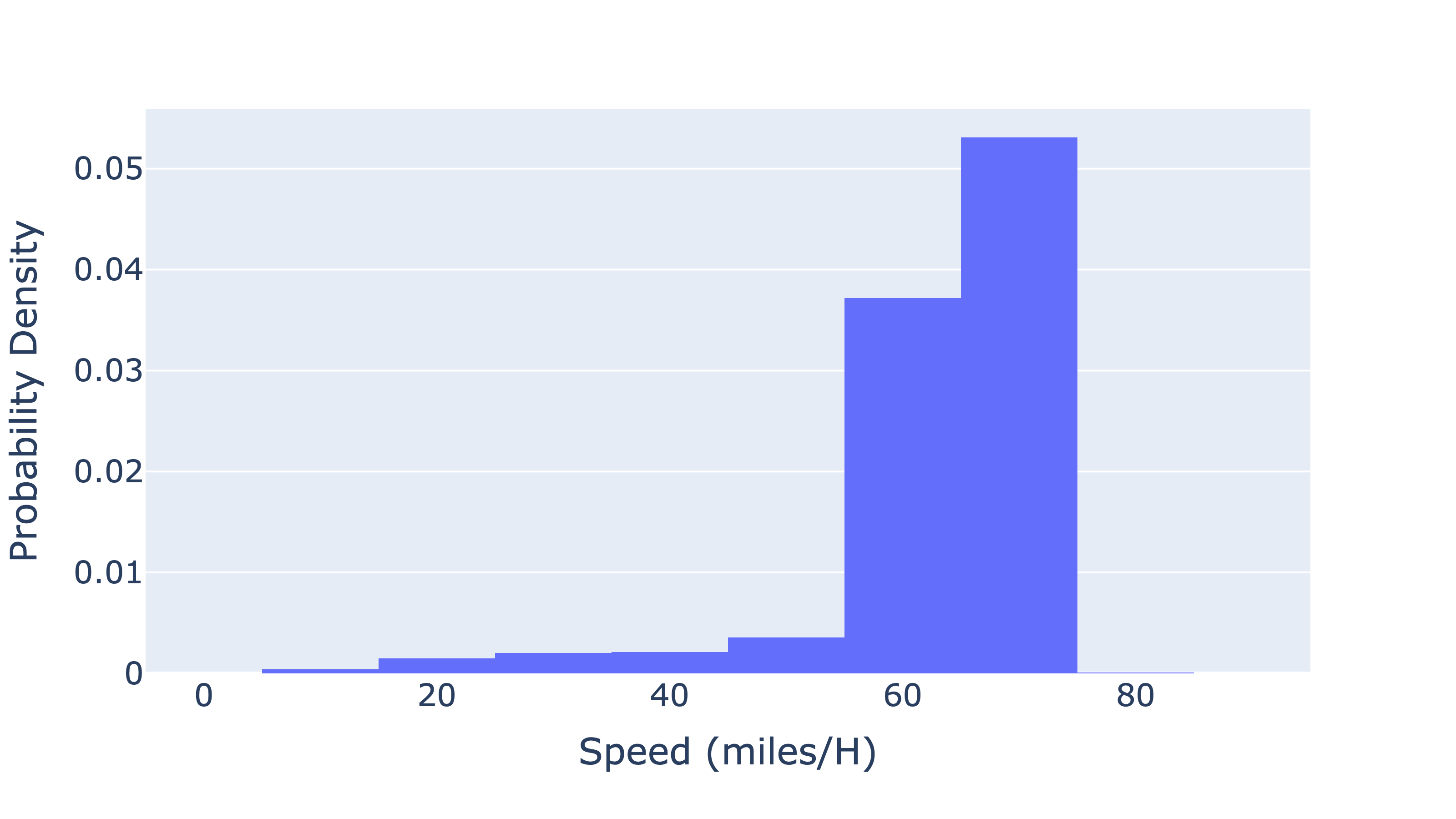}
        \caption{PEMS-BAY Speed Distribution}
        \label{fig:pems-speed}
    \end{subfigure}
    
    \vspace{0.05cm}
    
    \begin{subfigure}[b]{0.30\textwidth}
        \centering
        \includegraphics[width=\textwidth]{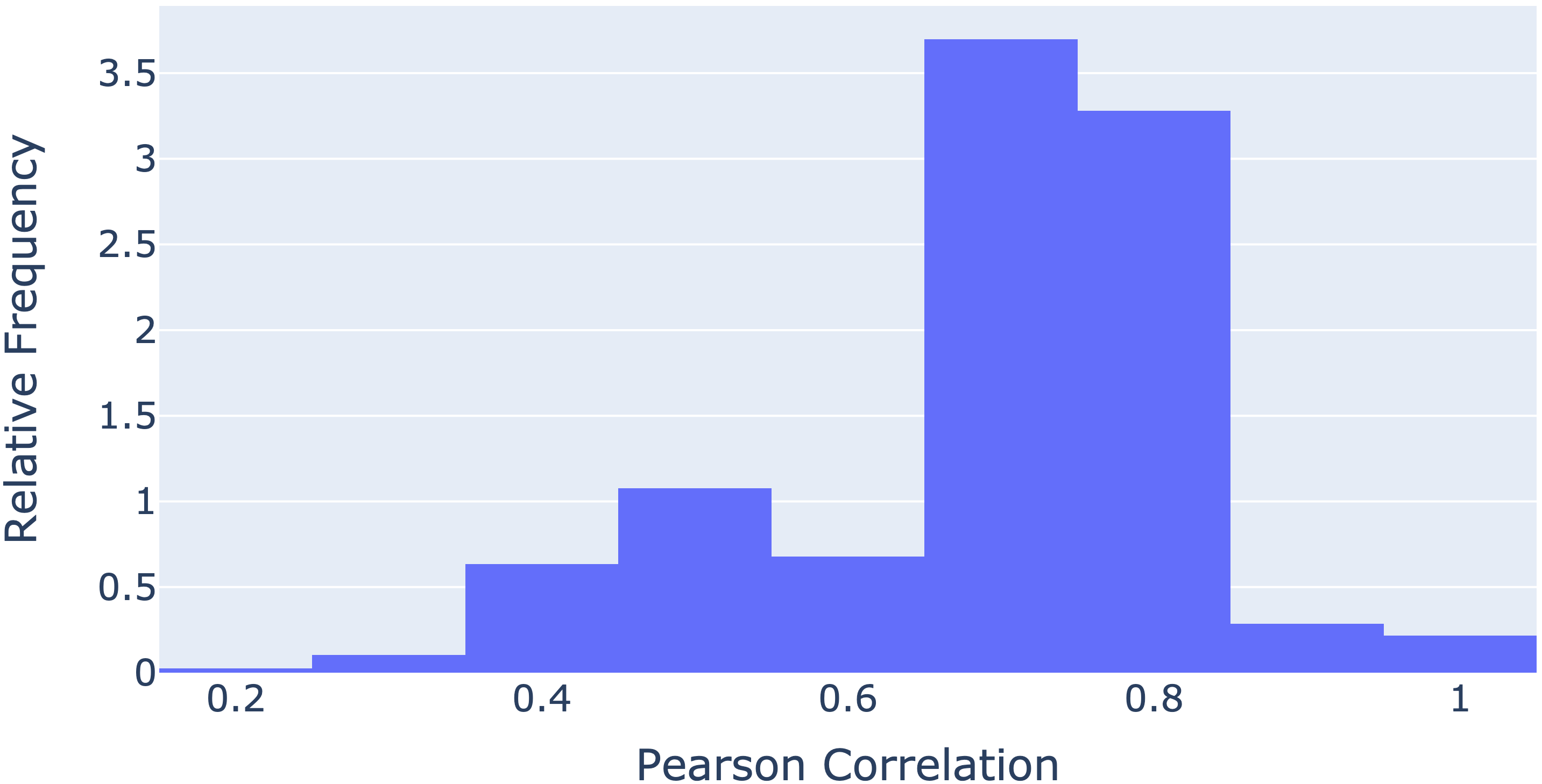}
        \caption{Hague Inter-node Correlation}
        \label{fig:hague-correlation}
    \end{subfigure}%
    \hfill
    \begin{subfigure}[b]{0.30\textwidth}
        \centering
        \includegraphics[width=\textwidth]{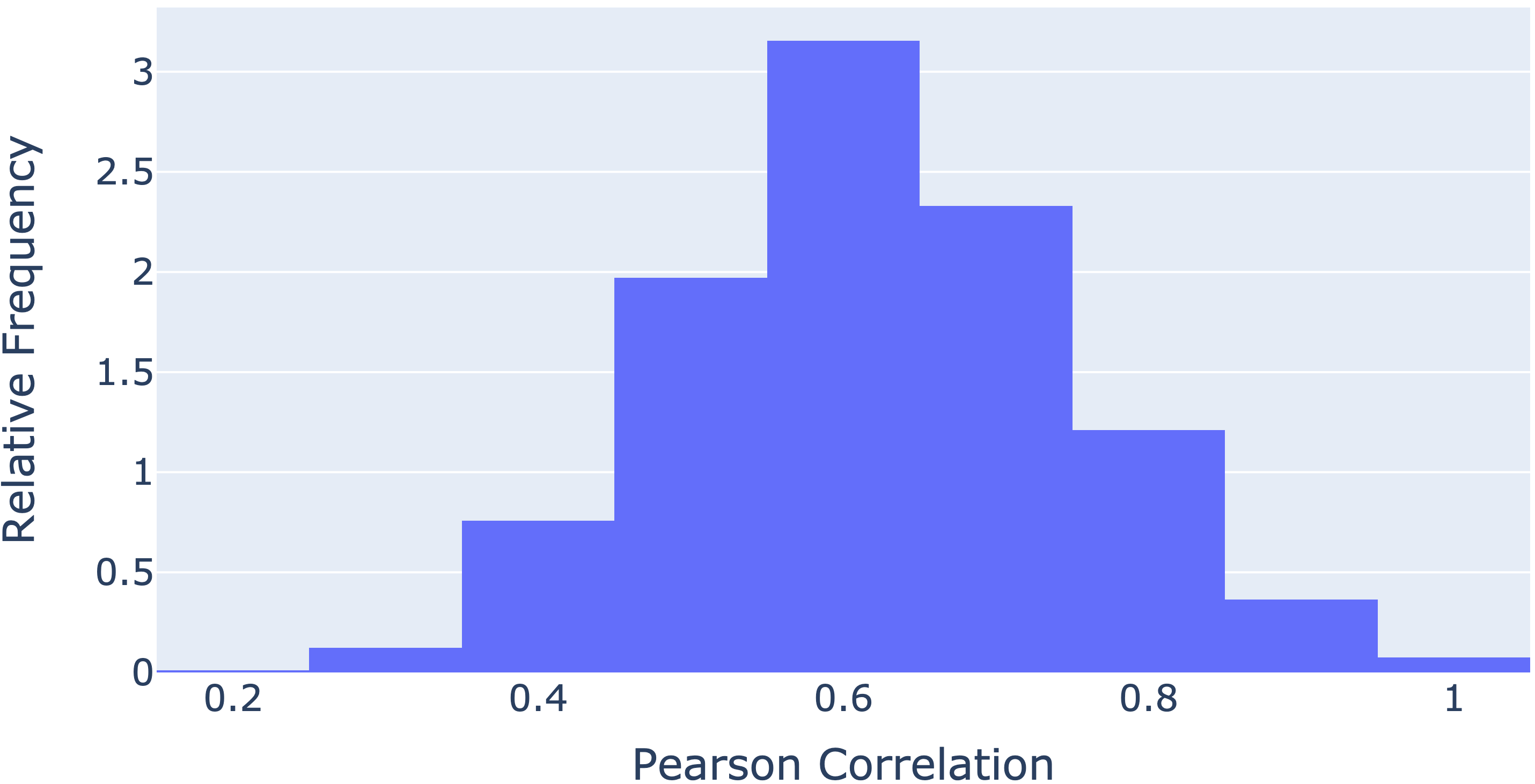}
        \caption{METR-LA Inter-node Correlation}
        \label{fig:metr-la-correlation}
    \end{subfigure}%
    \hfill
    \begin{subfigure}[b]{0.30\textwidth}
        \centering
        \includegraphics[width=\textwidth]{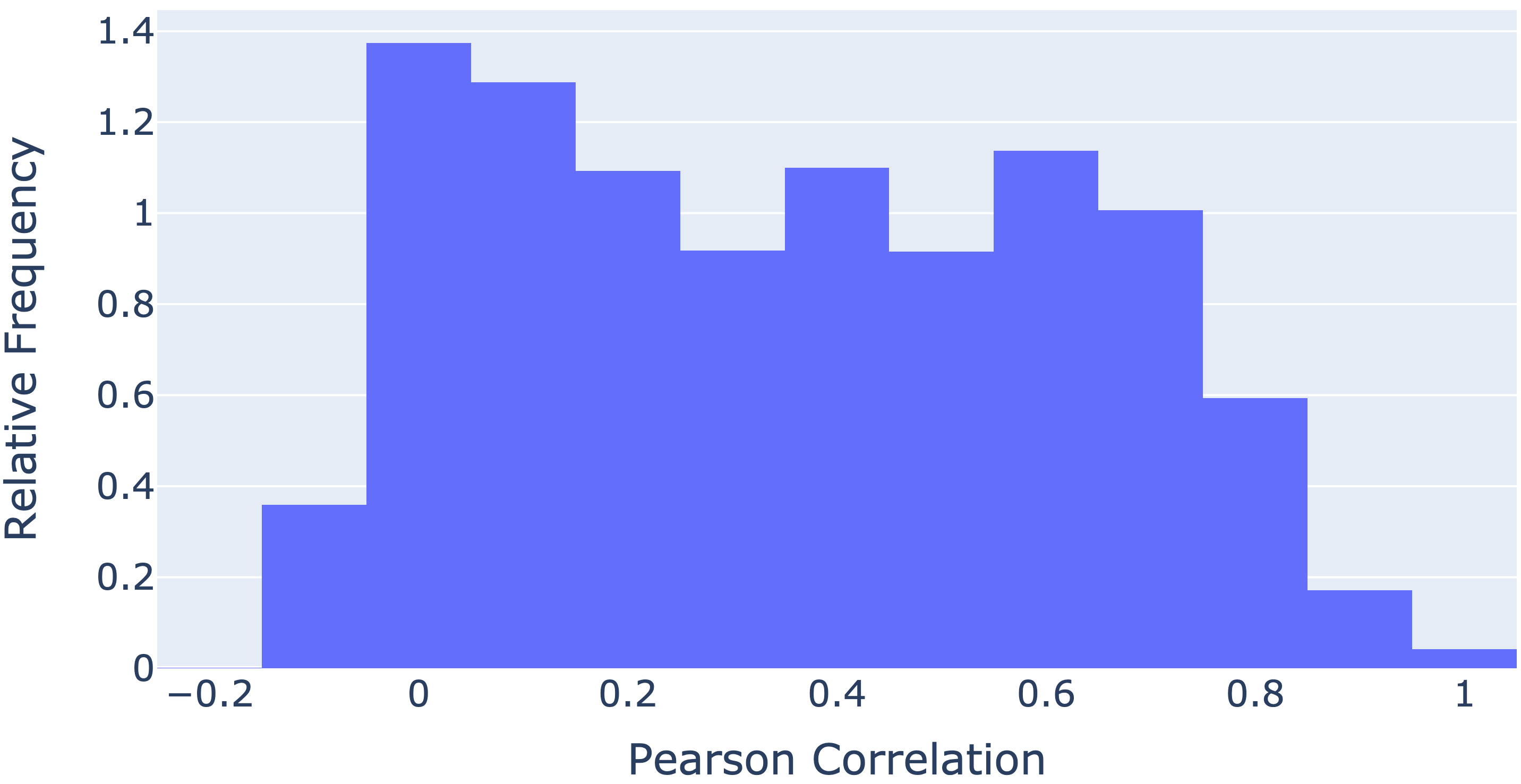}
        \caption{PEMS-BAY Inter-node Correlation}
        \label{fig:pems-correlation}
    \end{subfigure}
\end{figure*}

\subsubsection{PEMS-BAY Dataset}
\label{subsec:PemsBay}
Similar to the METR-LA dataset, the PEMS-BAY is also a widely used benchmark dataset for traffic modeling and forecasting. This public traffic speed dataset collected by California Transportation Agencies (CalTrans) contains 325 sensors in the Bay Area ranging from Jan 1st 2017 to May 31th 2017 denoting speed in miles/hour (Table \ref{tab:dataset-info}).

\begin{table}[ht]
\centering
\caption{Dataset Information}
\label{tab:dataset-info}
\begin{tabular}{|l|l|l|l|}
\hline
\textbf{Dataset} & \textbf{Hague} & \textbf{METR-LA} & \textbf{PEMS-BAY} \\
\hline
\textbf{\#Samples} & 236,736 & 34,272 & 52,116 \\
\textbf{\#Sensors} & 23 & 207 & 325 \\
\textbf{Sample Rate} & 5 minutes & 5 minutes & 5 minutes \\
\textbf{Time Range} & 2 Years & 4 months & 6 months \\
\textbf{Traffic Indicator} & \#Vehicles & Speed (mi/h) & Speed (mi/h) \\
\hline
\end{tabular}
\end{table}

In order to get higher-level insights into these datasets, we also computed the correlations among all pairs of sensors and examined their distributions. The results revealed significant inter-node (spatial) correlations in both METR-LA and Hague datasets, while the correlations in PEMS-BAY dataset are notably weaker. This observation is depicted in Figure \ref{fig:hague-correlation}, \ref{fig:metr-la-correlation} and \ref{fig:pems-correlation}.

By analyzing the velocity distributions presented in Figure \ref{fig:pems-speed}, \ref{fig:metr-la-speed}, and \ref{fig:hague-number}, notable differences can be observed among the datasets. PEMS-BAY dataset demonstrates a more uniform and closer-to-free-flow velocity distribution compared to METR-LA. This observation aligns with the lower inter-node correlations depicted in Figure \ref{fig:pems-correlation}, suggesting simpler and non-dependent traffic conditions in PEMS-BAY as compared to METR-LA. Furthermore, the vehicle distribution in Hague dataset reveals that most intersections experience low volumes of vehicles at any given time, indicating light and smooth traffic conditions in the area.

\subsection{System Details}

The experiments were performed on a Mac M1 machine with 16 GB RAM running on macOS 13. To leverage GPU capabilities, the experiments utilized the MPS backend. Python 3.10 served as the programming language for conducting the experiments. 

\section{RESULTS}
\label{sec:RESULTS}

To evaluate OWAM's performance comprehensively, we employed several criteria. Firstly, we investigated the impact of OWAM on the dimensionality of the traffic prediction model. Secondly, we examined the integration of Earth's Mover Distance loss in Autoencoders and its potential to enhance model predictions. Thirdly, we compared OWAM's performance with that of an LSTM Baseline and established state-of-the-art techniques. Lastly, we assessed OWAM's effectiveness in handling real-time scenarios, specifically its capability to adapt to dynamic traffic changes.

\begin{figure*}[htbp]
    \centering
    
    \begin{subfigure}[b]{0.49\linewidth}
        \includegraphics[width=\linewidth]{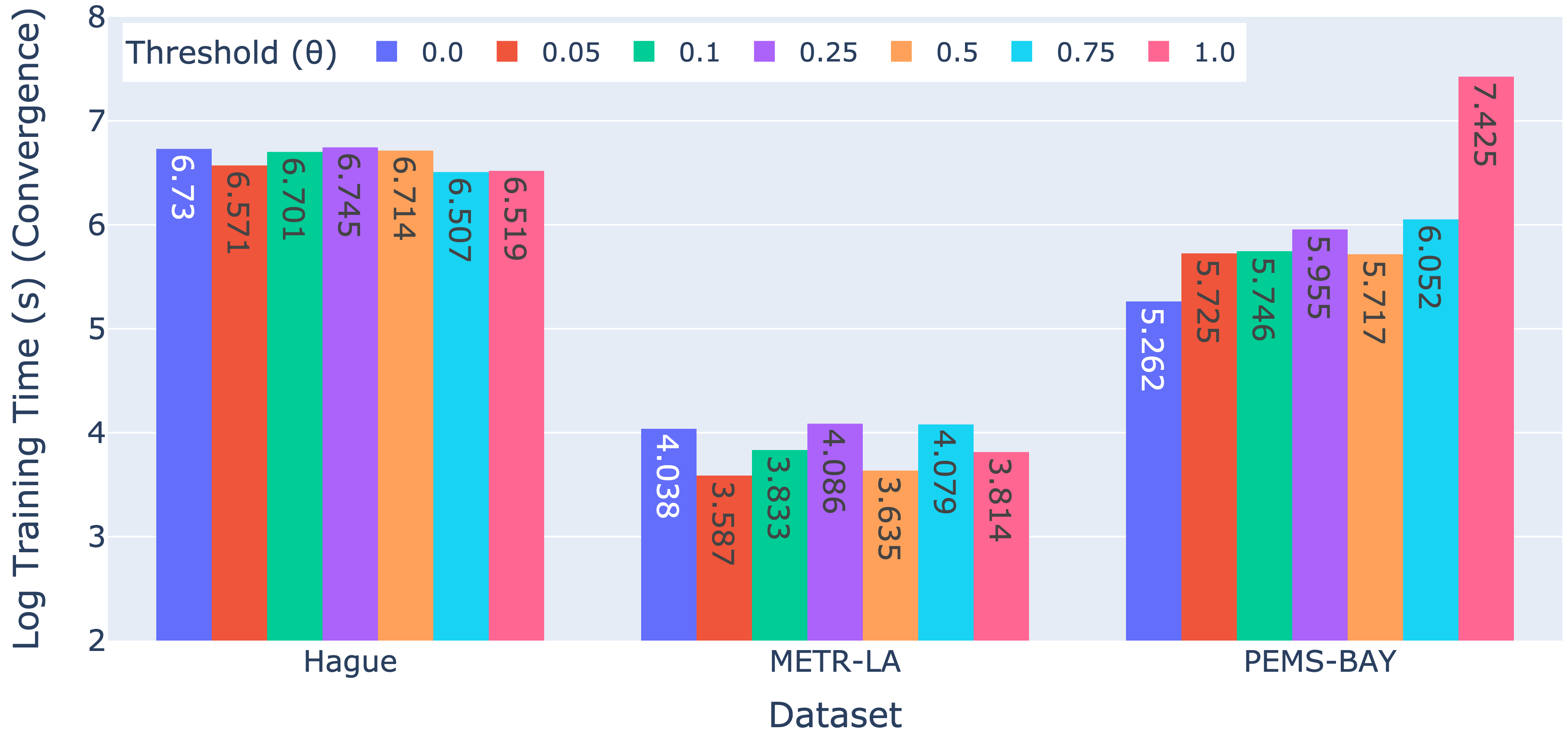}
        \caption{Model Training Time}
        \label{fig:training-time-optim}
    \end{subfigure}
    \hfill
    \begin{subfigure}[b]{0.49\linewidth}
        \includegraphics[width=\linewidth]{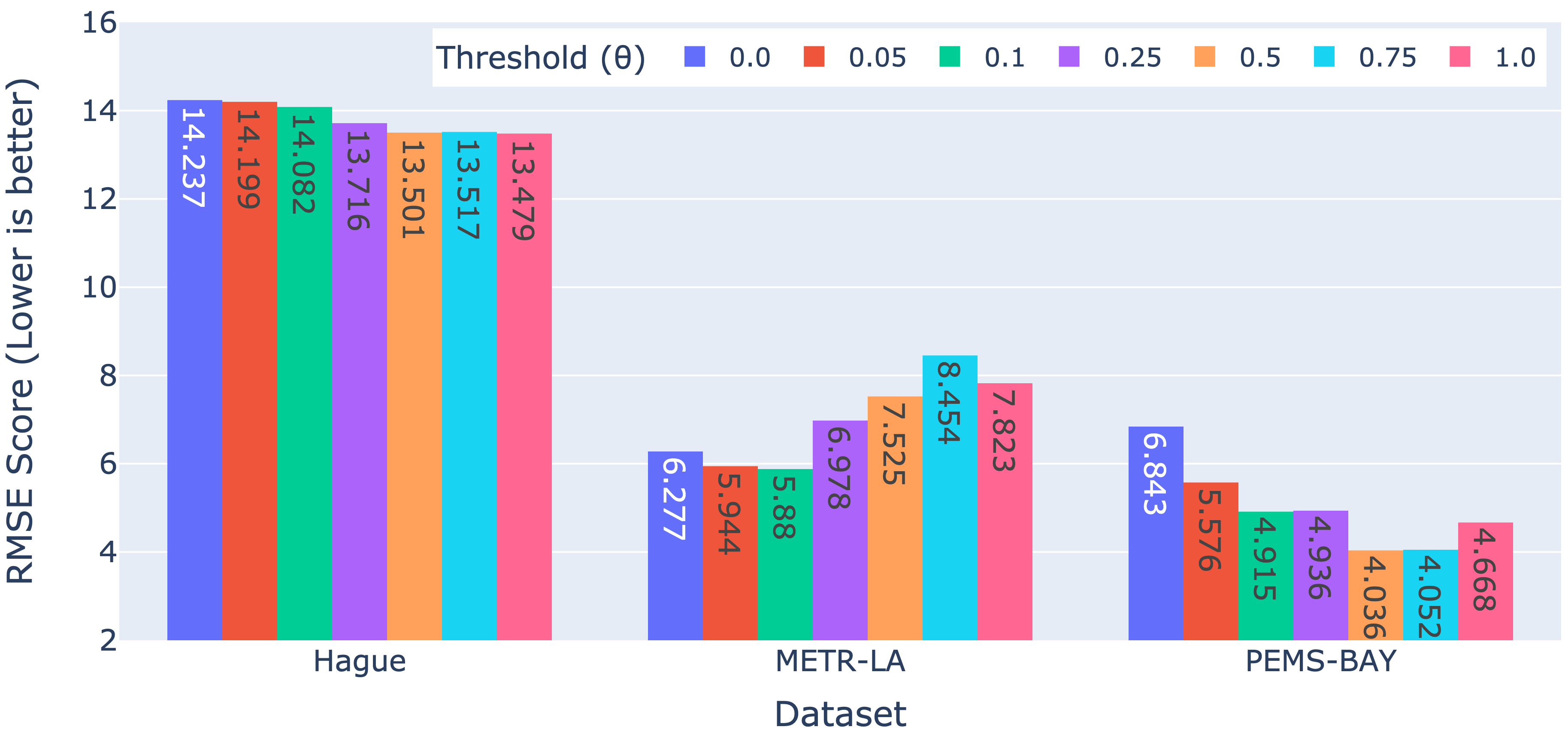}
        \caption{Model RMSE Score}
        \label{fig:rmse-optim}
    \end{subfigure}
    
    \caption{Model Training Time and RMSE Score for Different Thresholds for Hague, METR-LA, and PEMS-BAY Datasets}
    \label{fig:combined-optim-figures}
\end{figure*}

When considering dimensionality in traffic flow prediction, the challenge is to determine the optimal number of sensors to include in the model for predicting traffic flow at the target sensor. Figure \ref{fig:rmse-optim} illustrates the RMSE scores obtained from the LSTM-based OWAM framework for three datasets. Each dataset exhibits an optimal threshold value for achieving peak performance. The Hague dataset, comprising only 8 to 15 sensors, shows consistent performance when including the top 50\% or all neighboring sensors, making the dimensionality impact less pronounced. In contrast, METR-LA and PEMS-BAY datasets, which feature more sensors, experience performance degradation when considering all sensors. Instead, incorporating the top 10\% and 50\% of neighboring sensors yields improved predictions in these datasets. This highlights the dataset-specific nature of sensor selection.

It's worth noting that utilizing only the target sensor ($\theta = 0$) or incorporating all available sensor information ($\theta = 1$) results in lower prediction performance, as indicated by higher RMSE scores, across all datasets. This underscores the importance of selecting an optimal number of correlated sensors for both prediction accuracy and model efficiency, enabling the model to focus on influential factors while minimizing noise by including a limited subset of relevant sensors.

Regarding training time (Figure \ref{fig:training-time-optim}), there is no consistent trend observed across datasets. However, an increase in training time is noticeable for the higher-dimensional PEMS-BAY dataset. It is important to mention that we use model convergence time instead of total training time due to computational limitations, as this choice aligns with efficiency considerations.

\begin{table}[htbp]
\centering
\caption{Comparison of EMD Loss and RMSE Loss on Integration with Autoencoders}
\label{tab:model-results-grouped}
\begin{tabular}{|c|c|c|c|}
\hline
\textbf{Model} & \textbf{Dataset} & \textbf{Train Time} & \textbf{RMSE} \\
\hline
\multirow{2}{*}{OWAM (EMD Loss)} & Hague & 677.739 & \textbf{13.479} \\
 & METR-LA & \textbf{37.903} & \textbf{7.525} \\
 & PEMS-BAY & 313.028 & \textbf{4.915} \\
\hline
\multirow{2}{*}{OWAM (RMSE Loss)} & Hague & \textbf{245.465} & 13.539 \\
 & METR-LA & 81.914 & 7.918 \\
 & PEMS-BAY & \textbf{289.274} & 5.537 \\
\hline
\end{tabular}

\vspace{5pt}

\caption{Distribution of Outlier Score for RMSE and EMD Loss for a Random Sensor}
\label{tab:outlier_score_distribution}
\begin{tabular}{lcc}
    \hline
    & Mean & Standard Deviation \\
    \hline
    EMD Loss & 0.70 & 0.13 \\
    RMSE Loss & 0.81 & 0.11 \\
    \hline
\end{tabular}
\end{table}

Table \ref{tab:model-results-grouped} highlights the advantages of using Earth Mover's Distance (EMD) as the loss function. The OWAM model with EMD loss consistently outperforms the model relying on the RMSE loss (Euclidean Distance) function. While the improvements may not be substantial in the Hague scenario, an overall enhancement is still evident. The RMSE score decreases by 5\% and 11\% for METR-LA and PEMS-BAY, respectively. Although no clear pattern emerges in training times, it's important to note that RMSE is computationally more efficient, featuring O(n) time complexity, whereas EMD involves solving a linear programming problem, often less efficient than O(n), especially for large datasets.

To evaluate the advantages of incorporating EMD loss into the outlier model, refer to Table \ref{tab:outlier_score_distribution}. When EMD loss is used, it leads to a higher standard deviation in outlier scores compared to RMSE loss. This increased variance signifies the model's capacity to differentiate highly relevant sensors from less relevant ones, allowing it to accurately capture complex data patterns. Conversely, a lower standard deviation indicates less differentiation among sensors, potentially resulting in the oversight of important correlations and data patterns.

\begin{table}[htbp]
\centering
\caption{Results for the Hague dataset}
\label{tab:hague-results}
\begin{tabular}{cccccc}
\hline
Model Name & RMSE & Train Time & Instance Pred & Eval Time \\
  &  & (s) & Time (ms) & (s)  \\
\hline
OWAM & \textit{13.479} & 677.74 & \textbf{1.006} & \textbf{47.617} \\
HST\cite{HST} & 13.631 & \textbf{244.31} & 2.14 & 101.32 \\
Kit-Net\cite{Kitnet} & 13.540 & \textit{305.37} & \textit{1.019} & \textit{48.267} \\
DGCRN\cite{DGCRN} & \textbf{13.110} & 5043.31 & 5.91 & 280.00 \\
OBIS\cite{FPD-LOF} & 13.547 & 682.81 & 1.08 & 51.30 \\
LSTM Baseline & 13.547 & 682.81 & 0.63 & 29.83 \\
\hline
\end{tabular}
\end{table}

\begin{table}[htbp]
\centering
\caption{Results for the METR-LA dataset}
\label{tab:metr-results}
\begin{tabular}{cccccc}
\hline
Model Name & RMSE & Train Time & Instance Pred & Eval Time \\
  &  & (s) & Time (ms) & (s)  \\
\hline
OWAM & \textit{5.880} & \textbf{46.18} & \textbf{1.18} & \textbf{8.09} \\
HST\cite{HST} & 6.815 & \textit{47.53} & 3.99 & 27.37 \\
Kit-Net\cite{Kitnet} & 5.883 & 55.44 & 1.69 & 11.58 \\
DGCRN\cite{DGCRN} & \textbf{4.161} & 11263.23 & 45.23 & 310.04 \\
OBIS\cite{FPD-LOF} & 7.421 & 91.37 & \textit{1.57} & \textit{10.78} \\
LSTM Baseline & 8.831 & 106.06 & 1.62 & 11.11 \\
\hline
\end{tabular}
\end{table}

\begin{table}[htbp]
\centering
\caption{Results for the PEMS-BAY dataset}
\label{tab:pems-results}
\begin{tabular}{cccccc}
\hline
Model Name & RMSE & Train Time & Instance Pred & Eval Time \\
  &  & (s) & Time (ms) & (s)  \\
\hline
OWAM & \textit{4.036} & \textbf{303.94} & \textbf{14.23} & \textbf{97.56} \\
HST\cite{HST} & 4.816 & 372.86 & 16.15 & 110.73 \\
Kit-Net\cite{Kitnet} & 4.312 & \textit{356.96} & {15.23} & {104.393} \\
DGCRN\cite{DGCRN} & \textbf{1.591} & 25900.67 & 113.79 & 780.04 \\
OBIS\cite{FPD-LOF} & 4.375 & 377.49 & \textit{14.48} & \textit{99.27} \\
LSTM Baseline & 5.434 & 405.05 & 16.30 & 111.74 \\
\hline
\end{tabular}
\end{table}

Table \ref{tab:model-features} offers a comprehensive overview of the features associated with the compared approaches. The "LSTM Baseline" represents the standard modeling approach without any specific preprocessing steps. "DGCRN" (Diffusion Graph Convolution Recurrent Network) \cite{DGCRN} stands as a representative state-of-the-art graph-based approach in traffic analysis. The choice of DGCRN is driven by its recognition in the literature and the availability of open-source code, enabling valuable insights into the contributions of outlier-based research compared to cutting-edge graph-based techniques in traffic management. This comparison aids in assessing the practical potential of our framework in intelligent traffic management systems. "OBIS" \cite{FPD-LOF} represents the latest advancements in outlier-based approaches. Furthermore, we also compared OWAM to other online outlier-based methods (HST and Kit-Net), where sensors are selected similarly based on weighted correlation with only differentiation being the outlier technique. It should be noted that all state-of-the-art and other methods, including DGCRN, were applied using the same settings as OWAM to ensure a fair comparison. This approach guarantees the reproducibility of the results and provides a clear basis for evaluating OWAM's performance relative to other models.

\begin{table*}[htbp]
\centering
\caption{Feature Comparison of Traffic Prediction Models}
\label{tab:model-features}
\begin{tabular}{|c|c|c|c|c|c|}
\hline
\textbf{Model} & \textbf{Spatial Dependency} & \textbf{Temporal Dependency} & \textbf{Dimensionality Reduction} & \textbf{\shortstack{Neighboring Sensor\\Integration}} \\
\hline
LSTM Baseline & 
\begin{tabular}{c}
None
\end{tabular} & 
\begin{tabular}{c}
LSTM
\end{tabular} & 
\begin{tabular}{c}
No
\end{tabular} & 
\begin{tabular}{c}
Binary
\end{tabular} \\
\hline
DGCRN \cite{DGCRN} & 
\begin{tabular}{c}
\shortstack{Distance Matrix + \\ Graph Convolutions}
\end{tabular} & 
\begin{tabular}{c}
GRU
\end{tabular} & 
\begin{tabular}{c}
No
\end{tabular} & 
\begin{tabular}{c}
Binary
\end{tabular} \\
\hline
OBIS \cite{FPD-LOF} & 
\begin{tabular}{c}
\shortstack{Outlier Correlation \\ (LOF)}
\end{tabular} & 
\begin{tabular}{c}
LSTM
\end{tabular} & 
\begin{tabular}{c}
Yes
\end{tabular} & 
\begin{tabular}{c}
Binary
\end{tabular} \\
\hline
OWAM & 
\begin{tabular}{c}
\shortstack{Outlier Correlation \\ (AE)}
\end{tabular} & 
\begin{tabular}{c}
LSTM
\end{tabular} & 
\begin{tabular}{c}
Yes
\end{tabular} & 
\begin{tabular}{c}
Weighted
\end{tabular} \\
\hline
\end{tabular}
\end{table*}

Tables \ref{tab:hague-results}, \ref{tab:metr-results}, and \ref{tab:pems-results} provide summaries of the comparative analysis among different models. OWAM demonstrates superior performance compared to the previously developed outlier-based model and the Baseline LSTM in terms of RMSE scores. While the difference may not be highly pronounced for the Hague scenario, OWAM still enhances the average RMSE by 10\% compared to OBIS and 12\% compared to Baseline LSTM. Additionally, OWAM achieves similar or better training and evaluation times compared to OBIS and 15\% less training time than Baseline LSTM. Furthermore, OWAM surpasses both HST and Kit-Net in terms of RMSE and prediction time, attributed to its advanced outlier detection techniques and the weighted inclusion of top neighboring sensors. These enhancements contribute to improved prediction accuracy and computational efficiency.

OWAM delivers competitive performance to DGCRN in the Hague dataset, which boasts a relatively large sample size but fewer sensors. However, in the METR-LA and PEMS-BAY datasets, featuring significantly more sensors but relatively smaller sample sizes, DGCRN notably outperforms OWAM in terms of RMSE. This performance gap highlights DGCRN's prowess in handling complex traffic networks. DGCRN's superiority can be attributed to its effective utilization of spatial dependencies within the traffic network, leveraging sensor distances to construct adjacency matrices. In contrast, OWAM and the previous outlier-based model do not fully exploit sensor distances. However, DGCRN's advantages come at the cost of longer training times, as it requires a substantial amount of time to converge compared to OWAM and other models. Furthermore, Graph Neural Networks (GNNs) demand significantly higher memory consumption and a more extensive runtime than other models, making their prediction phase considerably longer. This limits their suitability for real-time traffic applications where prompt decisions are essential. 

The above comprehensive analysis indicates OWAM's superior performance compared to previous outlier-based and baseline techniques, coupled with significantly faster processing times than GNN-based techniques like DGCRN, making it the preferred choice for real-world applications. This speed is crucial for scalability and timeliness in large-scale, real-time traffic modeling, where rapid data processing and decision-making are essential. OWAM's ability to efficiently manage larger datasets and more complex scenarios with less computational burden than DGCRN positions it as a scalable, time-efficient solution for traffic management and similar real-time applications.

\begin{figure*}[htbp]
    \centering
    \begin{subfigure}[b]{0.32\linewidth}
        \includegraphics[width=\linewidth]{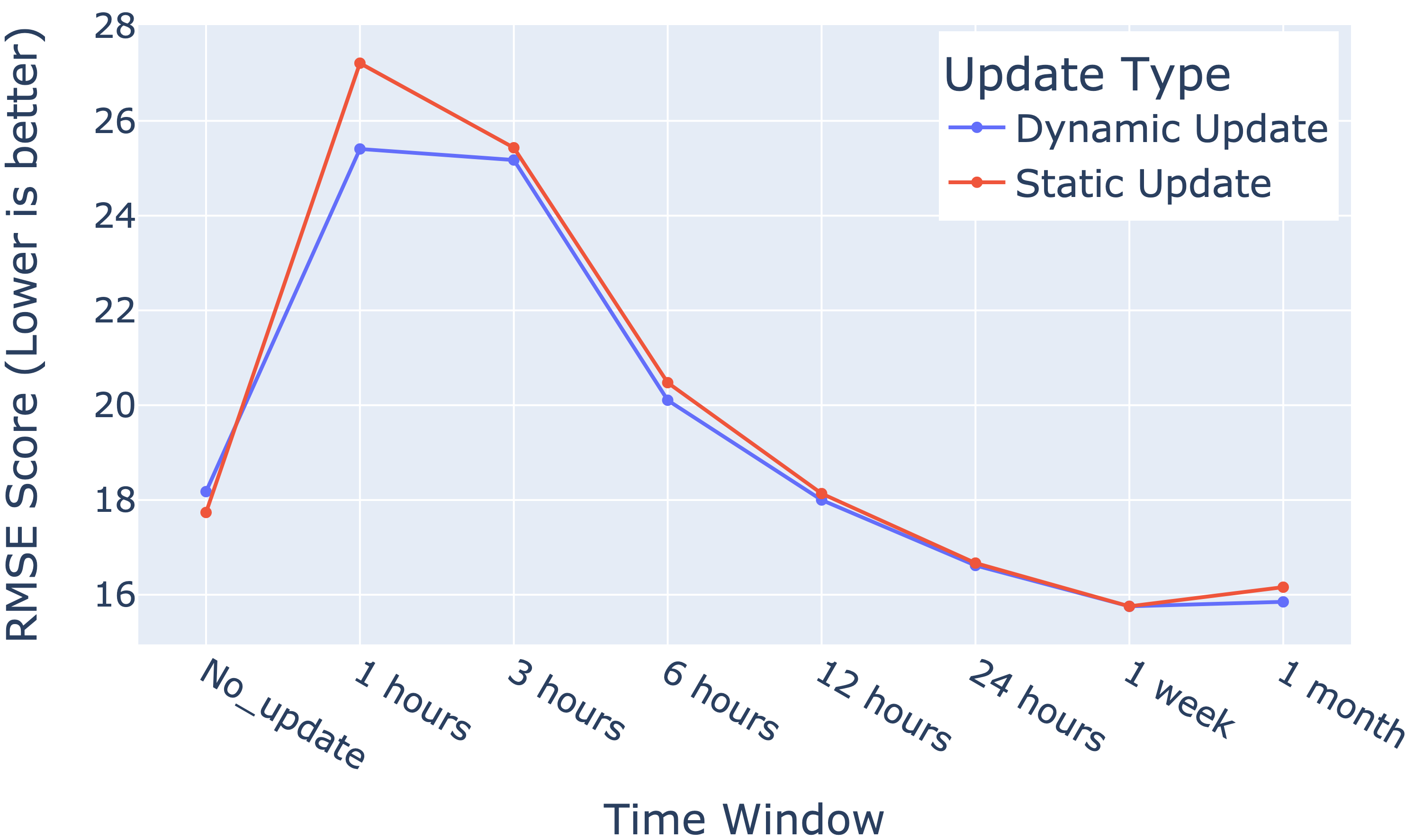}
        \caption{Hague Dataset}
        \label{fig:hague-rmse-rt}
    \end{subfigure}
    \hfill
    \begin{subfigure}[b]{0.32\linewidth}
        \includegraphics[width=\linewidth]{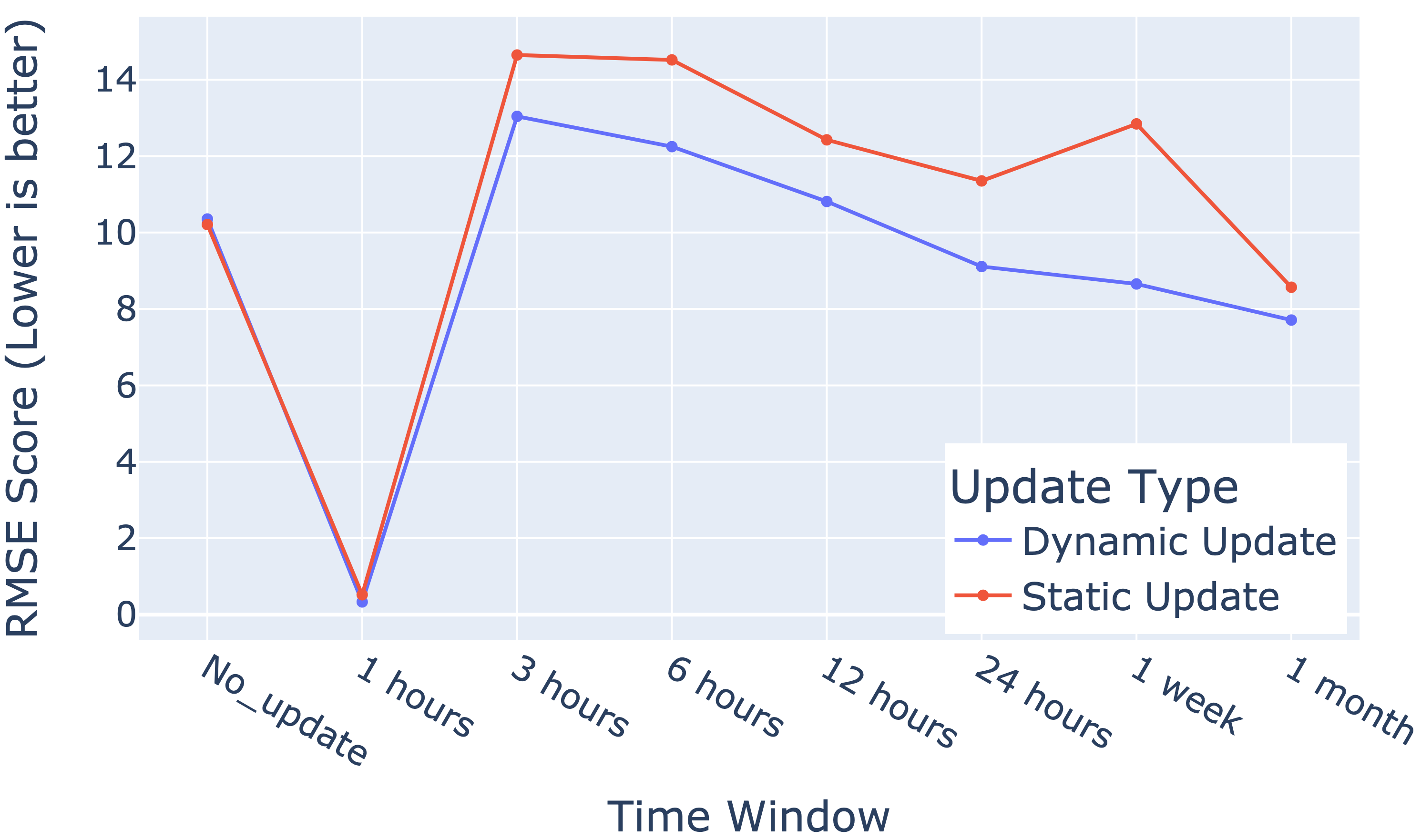}
        \caption{METR Dataset}
        \label{fig:metr-rmse-rt}
    \end{subfigure}
    \hfill
    \begin{subfigure}[b]{0.32\linewidth}
        \includegraphics[width=\linewidth]{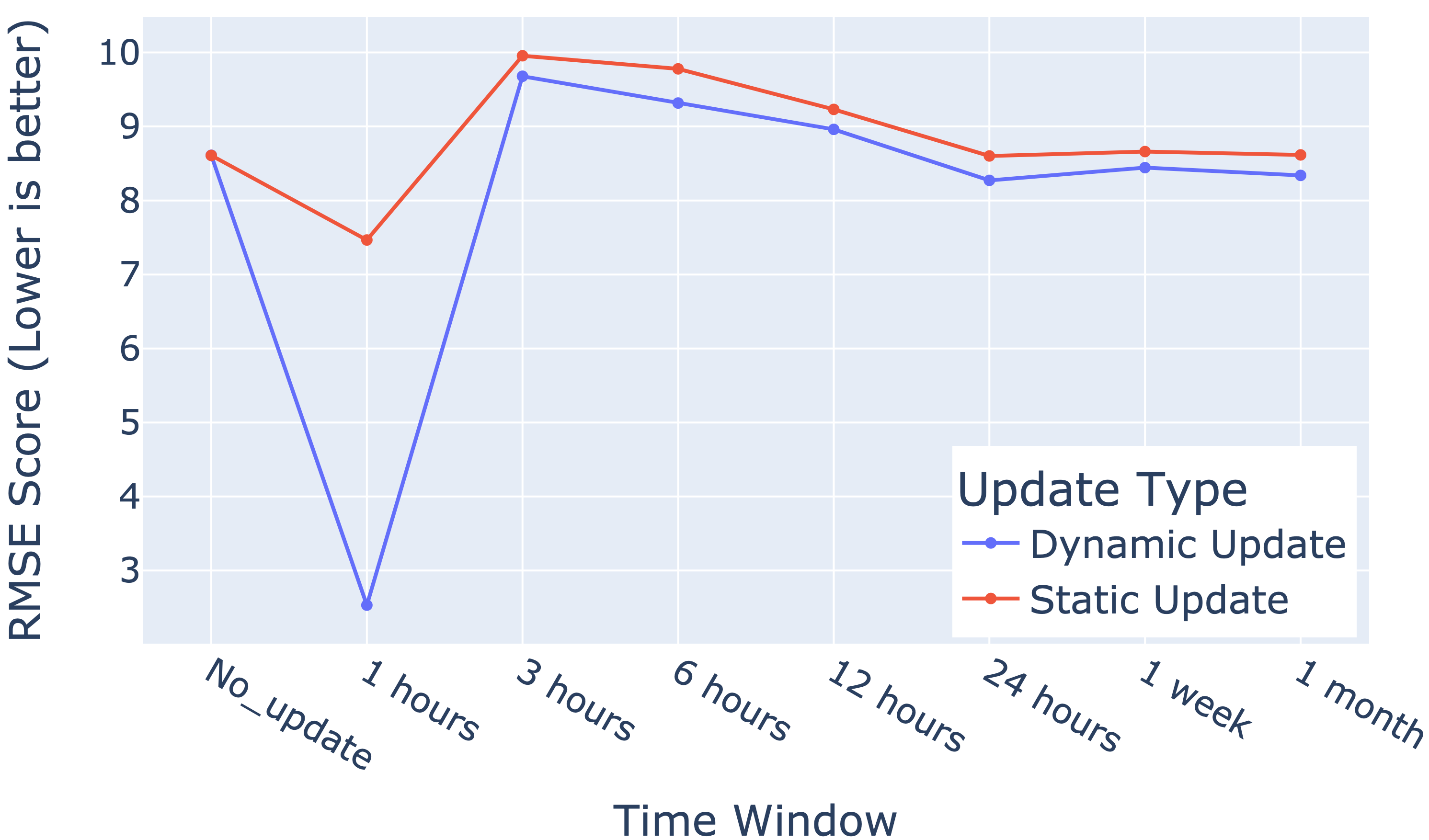}
        \caption{PEMS-BAY Dataset}
        \label{fig:pems-rmse-rt}
    \end{subfigure}
    \caption{Comparison Between Real-Time Dynamic and Static Updates over Different Datasets}
    \label{fig:real-time-comparisons}
\end{figure*}

Regarding real-time (online) settings, Figure \ref{fig:real-time-comparisons} demonstrates that dynamic updates based on outlier-weighted correlations (incremental updates using OWAM) significantly enhance the model's performance compared to static updates (incremental updates without OWAM, keeping neighboring sensors fixed) and no update models. This improvement is especially pronounced for very short (1-hour) or longer time windows (weekly or monthly), indicating their effectiveness in adapting to extended outlier situations and improving predictions. Contrasting results for the hague dataset at a 1-hour window indicate very small and frequent fluctuations in the traffic traffic.

The selection of an optimal update window is crucial in real-time modeling, influenced by specific requirements and computational constraints. While it might be assumed that smaller time windows always lead to better performance and improved model understanding due to more frequent updates, this is not universally true. In our experiments (Figure \ref{fig:real-time-comparisons}), we observed that updating the model every 3, 6, or 12 hours results in a decrease in performance compared to not updating it at all. This can be attributed to changes in traffic before even adapting to it. Therefore, it is essential to analyze the data and study the patterns before selecting a specific time window for model updates.

Overall, OWAM offers a promising solution for traffic flow prediction, demonstrating competitive performance when compared to baseline and state-of-the-art models. Its advanced outlier detection techniques and the weighted inclusion of top neighboring sensors contribute to enhanced accuracy and computational efficiency. However, the selection of the ideal model should be tailored to specific application requirements. Graph Neural Networks (GNNs) like DGCRN excel in accuracy-driven tasks, while OWAM provides a more efficient option for real-time applications.

\section{CONCLUSION AND FUTURE WORK}
\label{sec:CONCLUSION AND FUTURE WORK}

In conclusion, this paper introduces the Outlier Weighted Autoencoder Modeling (OWAM) framework, representing a pioneering approach to outlier-driven traffic modeling that addresses crucial gaps in the field. OWAM effectively integrates advanced outlier techniques into traffic modeling, intelligently incorporates outlier insights, and prioritizes real-time modeling. By combining autoencoders with Earth Mover's Distance loss, it enhances outlier detection and seamlessly incorporates anomaly information into traffic prediction models. Additionally, OWAM provides dynamic model updates for efficient real-time traffic modeling. Experimental findings showcase OWAM's remarkable ability to significantly improve prediction accuracy. It consistently outperforms other models, achieving an average RMSE score improvement of 10\% compared to OBIS and 12\% compared to the baseline LSTM model. While Graph Neural Networks (GNNs) demonstrate robust RMSE performance, OWAM excels in computational efficiency, positioning it as an attractive choice for real-time applications. The versatility and adaptability of OWAM make it a promising tool for real-world traffic analysis and management systems. Its contributions extend to the advancement of traffic modeling techniques and the enhancement of our understanding of urban traffic dynamics.

Future research should explore the use of more intricate datasets, such as NE-BJ \cite{DGCRN}, to better capture the complexities of urban traffic patterns. Conducting comparative analyses with a variety of traffic modeling approaches, including recent advanced Graph Neural Networks (GNNs) techniques \cite{Dynamic_Multi-View}, \cite{Dynamic_Correlation}, can further enhance our understanding of traffic prediction. Efforts to enhance the interpretability of GNNs are essential for making informed data-driven decisions in traffic management. The incorporation of elements from GNN architecture can be considered to achieve an equilibrium between the superior accuracy of GNNs and the rapid processing capabilities of outlier-based models. Additionally, merging weight allocation and model training into one unified network through the use of an attention mechanism could yield a robust and interpretable model. Moreover, OWAM's applicability extends to various domains that require anomaly detection and real-time responsiveness after immediate concept drift detection \cite{ICDM2016, ECMS2019, COMPSAC2023}, such as environmental monitoring, fraud detection \cite{Fastman}, and healthcare.

\bibliographystyle{ACM-Reference-Format}
\bibliography{references}  

\end{document}